\documentclass[conference]{IEEEtran}
\IEEEoverridecommandlockouts

\usepackage{cite}
\usepackage{amsmath,amssymb,amsfonts}
\usepackage{array}
\usepackage{algorithmic}
\usepackage{multirow}
\usepackage{multicol}
\usepackage{textcomp}
\usepackage{xcolor}
\usepackage{subfig}
\usepackage{graphicx, caption, subcaption}
\usepackage{adjustbox}
\usepackage{makecell}
\def\BibTeX{{\rm B\kern-.05em{\sc i\kern-.025em b}\kern-.08em
    T\kern-.1667em\lower.7ex\hbox{E}\kern-.125emX}}

\begin{document}

\title{Rasterized Steered Mixture of Experts for Efficient 2D Image Regression}

\author{Yi-Hsin~Li, Mårten~Sjöström,~\IEEEmembership{Senior Member,~IEEE,} Sebastian~Knorr,~\IEEEmembership{Senior Member,~IEEE,} \\Thomas Sikora,~\IEEEmembership{Senior Member,~IEEE}\vspace{-3ex}
\thanks{Manuscript received June 2025. This project has received funding from the European Union’s Horizon 2020 research and innovation program under the Marie Skłodowska-Curie grant agreement No 956770, and by Mid Sweden University internal funding. Computations were enabled by NAISS, partly funded by Swedish Research Council (2022-06725), and by High Performance Computing Center North (HPC2N) at Umeå University. \textit{(Corresponding authors: Mårten~Sjöström.)}}
\thanks{Yi-Hsin~Li is with Department of Computer and Electrical Engineering, Mid Sweden University, Sundsvall, 85170, Sweden, and also with Department of Telecommunication Systems, Technical University of Berlin, Berlin, 10587, Germany (e-mail: yi-hsin.li@miun.se).}
\thanks{Mårten~Sjöström is with Department of Computer and Electrical Engineering, Mid Sweden University, Sundsvall, 85170, Sweden (e-mail: Marten.Sjostrom@miun.se).}
\thanks{Sebastian~Knorr is with School of Computing, Communication and Business, Hochschule für Technik und Wirtschaft Berlin, Berlin, 12459, Germany (e-mail: sebastian.knorr@htw-berlin.de).}
\thanks{Thomas~Sikora is with Department of Telecommunication Systems, Technical University of Berlin, Berlin, 10587, Germany (e-mail: thomas.sikora@tu-berlin.de).}
}

\maketitle

\begin{abstract}
The Steered Mixture of Experts regression framework has demonstrated strong performance in image reconstruction, compression, denoising, and super-resolution. However, its high computational cost limits practical applications. {\color{black}This work introduces a rasterization-based optimization strategy that combines the efficiency of rasterized Gaussian kernel rendering with the edge-aware gating mechanism of the Steered Mixture of Experts.} The proposed method is designed to accelerate two-dimensional image regression while maintaining the model’s inherent sparsity and reconstruction quality. {\color{black}By replacing global iterative optimization with a rasterized formulation, the method achieves significantly faster parameter updates and more memory-efficient model representations.} In addition, the proposed framework supports applications such as native super-resolution and image denoising, which are not directly achievable with standard rasterized Gaussian kernel approaches. {\color{black}The combination of fast rasterized optimization with the edge-aware structure of the Steered Mixture of Experts provides a new balance between computational efficiency and reconstruction fidelity for two-dimensional image processing tasks.}
\end{abstract}






\section{Introduction}
\label{sec:introduction}
Our \textcolor{black}{primary} goal is to develop sparse regression models for efficient, fast, and high-quality modeling of images, fit for applications in inverse problems, such as image denoising. 

Image regression estimates a continuous function that maps image coordinates to pixel values. It enables smooth, high-fidelity representations of visual content and often employs neural networks or kernel-based models to enhance generalization and preserve fine details \cite{mildenhall2021nerf, salehi2022deep, Li_2023_CVPR, kerbl_3d_2023, zhang2024gaussianimage}.

In practice, image regression frequently involves reconstructing missing or corrupted data, casting it as an inverse problem. This problem is inherently ill-posed, requiring regularization or learned priors for stable recovery \cite{bertero2021introduction, hasanouglu2021introduction}. While classical approaches rely on handcrafted priors, modern methods employ deep networks to learn implicit structures \cite{genzel2022solving}, or leverage powerful \textcolor{black}{techniques} like plug-and-play priors \cite{zhang2021plug} and score-based diffusion models \cite{croitoru2023diffusion}.

Sparsity offers a powerful inductive bias \textcolor{black}{for addressing} ill-posedness. Sparse models reduce complexity, enhance interpretability, and support better generalization, especially under limited or noisy observations. Classical frameworks such as compressed sensing \cite{Baraniuk2010compressivesensing} and dictionary learning \cite{Mohamad2023dictionary} exploit sparsity for signal recovery, while recent advances embed similar ideas within deep architectures via sparse coding networks \cite{Yuli2022structured} and unrolled optimization schemes \cite{Kofler2023unrolling, Zhao2022imagefusionunrolling}.

\textcolor{black}{Recent works have advanced kernel-based methods for image regression and restoration by extending classical Gaussian models to deep or hybrid formulations. Quan et al.~\cite{quan2024gaussiandefocus} proposed a deep Gaussian kernel mixture for single-image defocus deblurring, demonstrating the ability of Gaussian mixtures to handle complex inverse problems while preserving fine structures. Li et al.~\cite{li2025adptsmoe} introduced an adaptive segmentation-based initialization strategy for steered mixture of experts and kernel regression, highlighting the importance of structured initialization in accelerating convergence and improving sparsity. Complementing these practical advances, Medvedev et al.~\cite{medvedev2025gaussianridgeless} analyzed the overfitting behavior and generalization bounds of ridgeless Gaussian kernel regression, offering theoretical insights relevant to sparse kernel-based regression frameworks. These studies underscore the continued relevance of Gaussian kernels and their variants for efficient and high-quality image regression, aligning closely with the principles motivating our SMoE approach.}

Motivated by the demands of sparse image regression, we focus on the Steered Mixture of Experts (SMoE) framework, a Gaussian-based model designed to efficiently handle sparse image regression through expert blending. SMoE's effectiveness has been demonstrated in 2D sparse image representation \cite{bochinski_regularized_2018}, 2D and 3D image and video compression \cite{jongebloed_sparse_2022}, and 4D/5D light field representation and coding \cite{verhack_steered_2020}. While steered Gaussian kernels dominate these models, strong performance has also been reported using steered Epanechnikov kernels, particularly for image and light field data \cite{liu_4d_2022}. SMoE has also shown promise in classic low-level vision applications: its edge-aware formulation has recently been leveraged for image denoising and restoration, yielding competitive results in a domain that has seen sustained attention over the past two decades \cite{ozkan_steered_2023, fleig_steered_2023}. Overall, SMoE presents a versatile and principled approach used for denoising, super-resolution, and high-dimensional data representation.

SMoE combines sparsity with edge-awareness, making it especially well-suited for inverse problems. At the core of the SMoE model is an edge-aware kernel representation, which adapts to the local structures in images, preserving crucial details while performing operations such as sparse representation for regression and compression, noise reduction, and resolution enhancement. Sparsity and edge-preservation in SMoE regression models are achieved using a \textcolor{black}{gating} network, i.e. with normalized steered Gaussian kernels. Well-known kernel regression frameworks, such as Radial Basis Function networks in Gaussian Splatting (GS-RBF)  \cite{kerbl_3d_2023} or Takeda Kernel Regression \cite{takeda2007kernelregression}, model pixel values as ``weighted sum of kernels''. In contrast, the SMoE gating network models pixels as ``weighted sum of gating functions'' using strategies similar to \textcolor{black}{\textit{normalized RBF} \cite{Heimes1998nrbf}} networks. This ensures \textcolor{black}{the} sparsity of the SMoE model -- few kernels are sufficient to represent complex texture\textcolor{black}{s}. An optimization process ensures that SMoE \textcolor{black}{\textit{gating functions}} align with edges in images easily and efficiently. Sharp edges and smooth transitions in images are represented with sparse SMoE models using few kernels. \textcolor{black}{By comparison,} RBF regression frameworks, on the other hand, usually require many kernel functions for edge reconstruction. 

Despite \textcolor{black}{SMoE's} sparsity and edge-awareness, optimizing \textcolor{black}{these models} remains a central challenge. The high parameter count and non-convex objective landscape make gradient descent \textcolor{black}{slow and computationally expensive}. For each image, a unique set of optimal parameters \textcolor{black}{must} be identified. As with RBF network optimization, SMoE optimization \textcolor{black}{typically} requires minimizing \textcolor{black}{highly} non-convex objective functions with thousands of parameters. This can result in excessive runtimes, \textcolor{black}{rendering many SMoE approaches impractical for real-world deployment}. There are primarily two strategies \textcolor{black}{for implementing} SMoE models for regression: (a) block-wise SMoE \cite{tok_mse_2018, fleig_edge-aware_2022}, which optimizes kernels locally using either autoencoders or iterative solvers, and (b) global SMoE \cite{bochinski_regularized_2018,jongebloed_hierarchical_2018,jongebloed_quantized_2019,jongebloed_sparse_2022}, which jointly optimizes all kernels over the full image. While block-wise SMoE offers speed, its local view can compromise quality. Global SMoE delivers greater sparsity and fidelity, but at the cost of significant computational effort.

Gaussian Splatting (GS) \cite{kerbl_3d_2023, zhang2024gaussianimage}, a recent method originally designed for 3D scene representation and rendering using volumetric Gaussian kernels—\textcolor{black}{and conceptually related to earlier surface splatting techniques} \cite{Zwicker2001surfacesplatting}, shares several \textcolor{black}{similarities } with SMoE. Both methods splat \textcolor{black}{s}teered Gaussians into the pixel \textcolor{black}{domain} and \textcolor{black}{use} gradient descent to optimize Gaussian kernels \textcolor{black}{for effective data representation}. GS leverages localized block-rasterization to accelerate training, primarily targeting 3D applications. The key distinction between GS and SMoE lies \textcolor{black}{in the GS-RBF regression model versus the SMoE normalized gating network} for regression. SMoE regression can be seen as an extended, more powerful strategy \textcolor{black}{compared} to GS-RBF with normalization using the same kernel representation.

First attempts to adopt the rasterized GS strategy to 2D image regression appeared in \cite{zhang2024gaussianimage}. \textcolor{black}{Building decisively on this foundation, our work is the first to integrate GS-inspired rasterization directly into the SMoE framework, combining GS’s computational speed with SMoE’s superior sparsity and edge-aware reconstruction.} The purpose of this paper is to take advantage of both GS (fast) and SMoE (sparse and high-quality reconstruction) approaches to arrive at a fast and sparse high-quality SMoE regression method. To this end, we seek to adopt the rasterization approach of GS to optimize SMoE parameters, \textcolor{black}{overcoming the critical bottleneck of slow, iterative gating network optimization and enabling drastic run-time improvements.}

The main contributions of our work are summarized as follows:

\begin{itemize}
\item We provide insights into the different functioning and capabilities of RBF (Gaussian Splatting) and SMoE regression frameworks, \textcolor{black}{with particular emphasis on SMoE’s superior edge reconstruction, denoising, and sharpening properties}.
\item We evaluate non-rasterized RBF 2D regression and SMoE regression performance with global optimization and \textcolor{black}{clearly demonstrate the superior sparsity and reconstruction quality of SMoE}.
\item We introduce R-SMoE, a rasterized SMoE training framework that leverages tile-based rasterization inspired by Gaussian Splatting. Compared to non-rasterized SMoE regression, R-SMoE accelerates model training and rendering times by orders of magnitude \textcolor{black}{while maintaining high fidelity}. Compared to “GaussianImage” \cite{zhang2024gaussianimage}, which employs GS-RBF rasterization, \textcolor{black}{our approach} requires significantly fewer computational resources for both training and rendering.
\item We introduce a segmentation-guided multi-hypothesis strategy tailored for denoising, enhancing the performance of both R-SMoE and GaussianImage by leveraging structural priors during inference.
\item We provide a systematic comparison between global SMoE and RBF optimization versus their rasterized counterparts—R-SMoE and GS-RBF—highlighting efficiency-quality trade-offs \textcolor{black}{and positioning R-SMoE as a practical, resource-efficient solution for high-quality image regression and denoising}.
\end{itemize}

\begin{figure}
    \centering
    \subfloat[Original]{\includegraphics[width=0.27\linewidth]{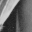}}
    \hfill
    \subfloat[SMoE Kernels]{\includegraphics[width=0.27\linewidth]{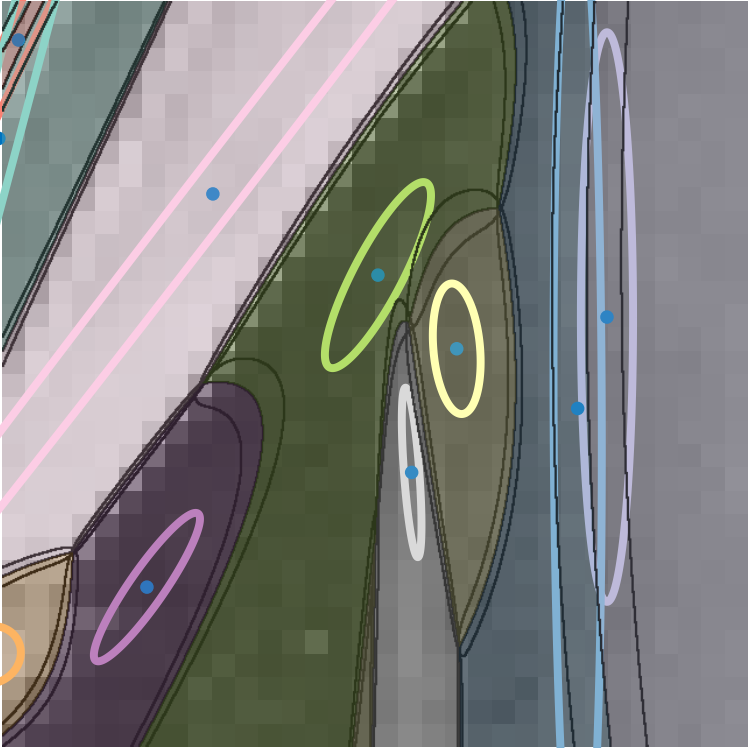}}
    \hfill
    \subfloat[SMoE Gates]{\includegraphics[width=0.27\linewidth]{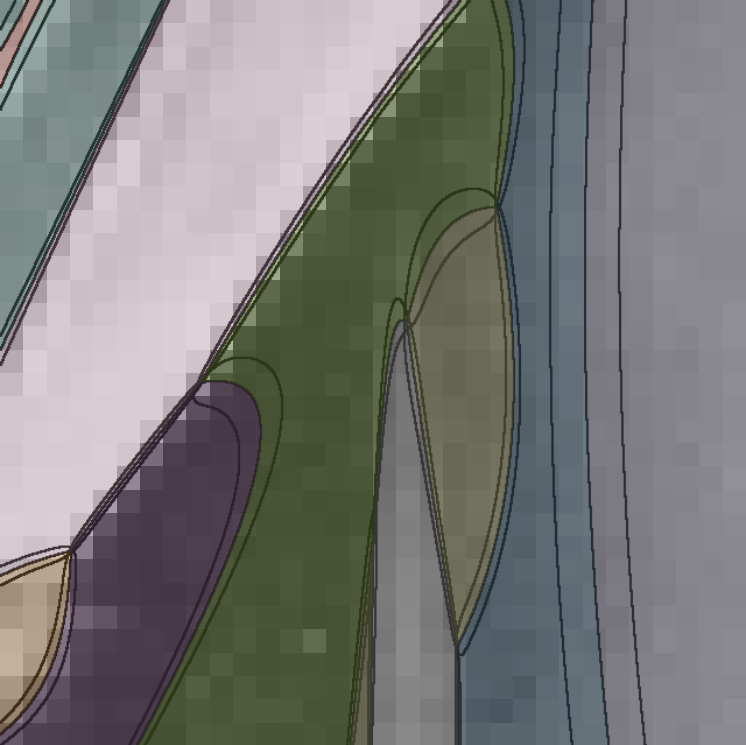}}
    \hfill
    \subfloat[\textbf{JPEG}\\ PSNR:$26.33 dB$\\ SSIM: $0.82$]{\includegraphics[width=0.24\linewidth]{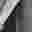}}
    \hfill
    \subfloat[\textbf{HEVC}\\PSNR:$26.05 dB$ \\SSIM: $0.77$]{\includegraphics[width=0.24\linewidth]{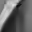}}
    \hfill
    \subfloat[\textbf{JPEG2000}\\ PSNR:$29.43 dB$\\ SSIM: $0.87$]{\includegraphics[width=0.24\linewidth]{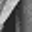}}
    \hfill
    \subfloat[\textbf{SMoE}\\ PSNR:$31.66 dB$ \\SSIM: $0.9$]{\includegraphics[width=0.24\linewidth]{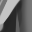}}
\vspace{-0.15cm}
\caption{Illustration of the edge-aware Steered Mixture of Experts (SMoE) \textcolor{black}{model applied to image} compression and denoising at \textcolor{black}{0.43 bits per pixel (bpp)}. \textcolor{black}{Figure adapted from} \cite{fleig_edge-aware_2022}.}
\label{fig1}
\vspace{-0.4cm}
\end{figure}

\section{Steered Mixture of Experts (SMoE)}
The SMoE image model describes an edge-aware, parametric, continuous \textcolor{black}{nonlinear} regression function. Fig. \ref{fig1} (from \cite{fleig_edge-aware_2022}) illustrates the soft-gating concept of the kernel model for image compression and denoising tasks.

SMoE gating networks can explicitly model and reconstruct both sharp and smooth transitions in images without straddling edges. The sparse, edge-aware SMoE model can reconstruct the original pixel block with excellent edge quality. Unlike traditional compression schemes like JPEG, JPEG2000, and HEVC-Intra, which often produce \textcolor{black}{visible} blocking \textcolor{black}{or} ringing artifacts at \textcolor{black}{similar} bit rate\textcolor{black}{s}, SMoE avoids these issues entirely.

Unlike traditional compression schemes like \textcolor{black}{JPEG and HEVC-Intra}, which often produce blocking and ringing artifacts at the same bit rate, SMoE avoids these issues entirely.  \textcolor{black}{JPEG2000, while free from blocking due to its wavelet-based design, may still introduce noise-like distortions at low bit rates. In contrast, SMoE delivers artifact-free reconstructions with higher visual fidelity.}

JPEG-like compression schemes operate in the \textcolor{black}{\textit{frequency}} domain, quantizing and coding DCT- or wavelet-coefficients, leading to \textcolor{black}{\textit{ringing}} artifacts at low rates. In contrast, SMoE models perform compression in the \textcolor{black}{\textit{pixel}} domain by quantizing and coding kernel parameters. \textcolor{black}{JPEG-like compression} results in geometric distortions of edges and lines at low rates, which are not directly visible in the reconstructed SMoE image. Both objective and subjective quality measures are greatly enhanced. \textcolor{black}{While SMoE excels at preserving sharp structural boundaries, it may produce slight blurring in highly stochastic textures, such as fur or grass; for detailed discussion and illustrative failure cases, see Section~VIII.}

SMoE regression employs Gaussian \textcolor{black}{steered} kernels distributed across multiple grids of pixels. The position and steering parameters of these kernels are optimized using gradient descent (GD) optimization for each image or block. Several  \textcolor{black}{\textit{experts}} collaborate to explain the data in specific 2D image regions, while the associated 2D soft-gating functions define the actual influence of each expert on each pixel. The ``gates,'' represented by 2D softmax functions, define boundary transitions in images, such as sharp edges and smooth transitions—providing the edge-awareness of the model. Sharp edges are modeled with sharp gating functions, while smooth transitions are modeled using overlapping gates. In Fig. \ref{fig1}, the sparse SMoE model with only 10 kernels is sufficient to explain the significant pixel variations in the image, while simultaneously  \textcolor{black}{\textit{denoising}} the pixels.

\subsection{Theory}
SMoE regression attempts to fit the image data using a combination of kernels. Each kernel's parameters (center $\mu$ and variance $\Sigma$) and experts are jointly adjusted during optimization to minimize the error between the reconstructed image and the original. The number of kernels is a critical factor in optimization time because of the complexity and sparsity considerations. More kernels increase the complexity of the regression function, leading to more parameters that need to be optimized. This directly impacts the computational burden and time required for convergence. Furthermore, each kernel interacts with multiple pixels, and the optimization process must account for these interactions. More kernels mean more calculations per iteration, significantly increasing the time required for each gradient descent step. \textcolor{black}{They} also mean higher memory consumption, as each kernel's parameters must be stored and updated during optimization. As a result, finding the balance between the number of kernels and the optimization time is crucial. Too few kernels might not capture the image details accurately, while too many can lead to excessive computational demands without significant quality improvement.

The Steered Mixture of Experts (SMoE) model extends traditional kernel regression methods, such as Radial Basis Function (RBF) \cite{ghosh_overview_2001} networks used in Gaussian Splatting, by introducing a more sophisticated mechanism for image reconstruction. \textcolor{black}{Although Gaussian Splatting also uses a “weighted sum of Gaussians,” which is conceptually similar to RBF regression, the interpretation differs: in RBF, the weights represent the importance of each Gaussian basis, whereas in Gaussian Splatting, the weights correspond to the color (appearance) of the Gaussian. Thus, while the regression form is similar, the modeling objectives are fundamentally different.} RBF networks  define the regression function \( y_p(x) \) based on a weighted sum of \( L \) (steered 2D Gaussian) kernels \( K_j(x) \):

\begin{equation}
y_p(x) = \sum_{j=1}^{L} m_j \cdot K_j(x)
\label{eq1}
\end{equation}
with steered 2D Gaussian kernels
\begin{equation}
K_j(x) = \exp\left(-\frac{1}{2} (x - \mu_j)^\text{T} \Sigma_j^{-1} (x - \mu_j)\right).
\label{eq3}
\end{equation}
In these equations, \( \mu_j \) and \( \Sigma_j \) represent the 2D center vectors and 2x2 covariance matrices of the steered Gaussian kernels, respectively. 
 \( x \) represents the pixel coordinates in the 2-dimensional continuous signal space over which the pixels are defined, and \( y_p(x) \) the estimated pixel amplitude at coordinate \( x \). \( m_j \) is the kernel weight.

In contrast, the SMoE model leverages a gating network approach, operating on a weighted sum of \( L \) 2D soft-gates \( w_j(x) \). The SMoE regression function is defined as:
\begin{equation}
y_p(x) = \sum_{j=1}^{L} m_j(x) \cdot w_j(x)
\label{eq2}
\end{equation}
The associated 2D soft-gating functions \( w_j(x) \) are derived by the softmax function:
\begin{equation}
w_j(x) = \frac{\pi_j \cdot K_j(x)}{\sum_{i=1}^{L} \pi_i \cdot K_i(x)}
\label{eq4}
\end{equation}

The \(\pi_i\) \textcolor{black}{values} provides additional weights to each kernel. Expert functions \( m_j(x) \) can take various functional forms, including constant, linear, quadratic, and basis functions such as DCT and wavelet bases \cite{blu2002wavelets}. For our work, we utilize simple constant experts \( m_j(x) = m_j \), which have demonstrated excellent performance in previous studies \textcolor{black}{\cite{bochinski_regularized_2018, tok_mse_2018}}. The irregularly shaped 2D softmax gating functions \( w_j(x) \) are derived based on the position and parameters of the \( L \) kernels, allowing for precise modeling of both smooth transitions and sharp edges in images.

\textcolor{black}{We note that the primary distinction between RBFs and SMoEs lies in the use of a weighted sum of kernels versus a weighted sum of soft-gates. Each soft-gate represents a normalized kernel, interpreted as a conditional distribution of Gaussians, which aligns with the neural network paradigm and enables precise modeling of boundaries—capabilities that cannot be achieved with pure Gaussian kernels alone.}

\begin{figure}
    \centering
    \subfloat[Denoising]{\includegraphics[width=0.49\linewidth]{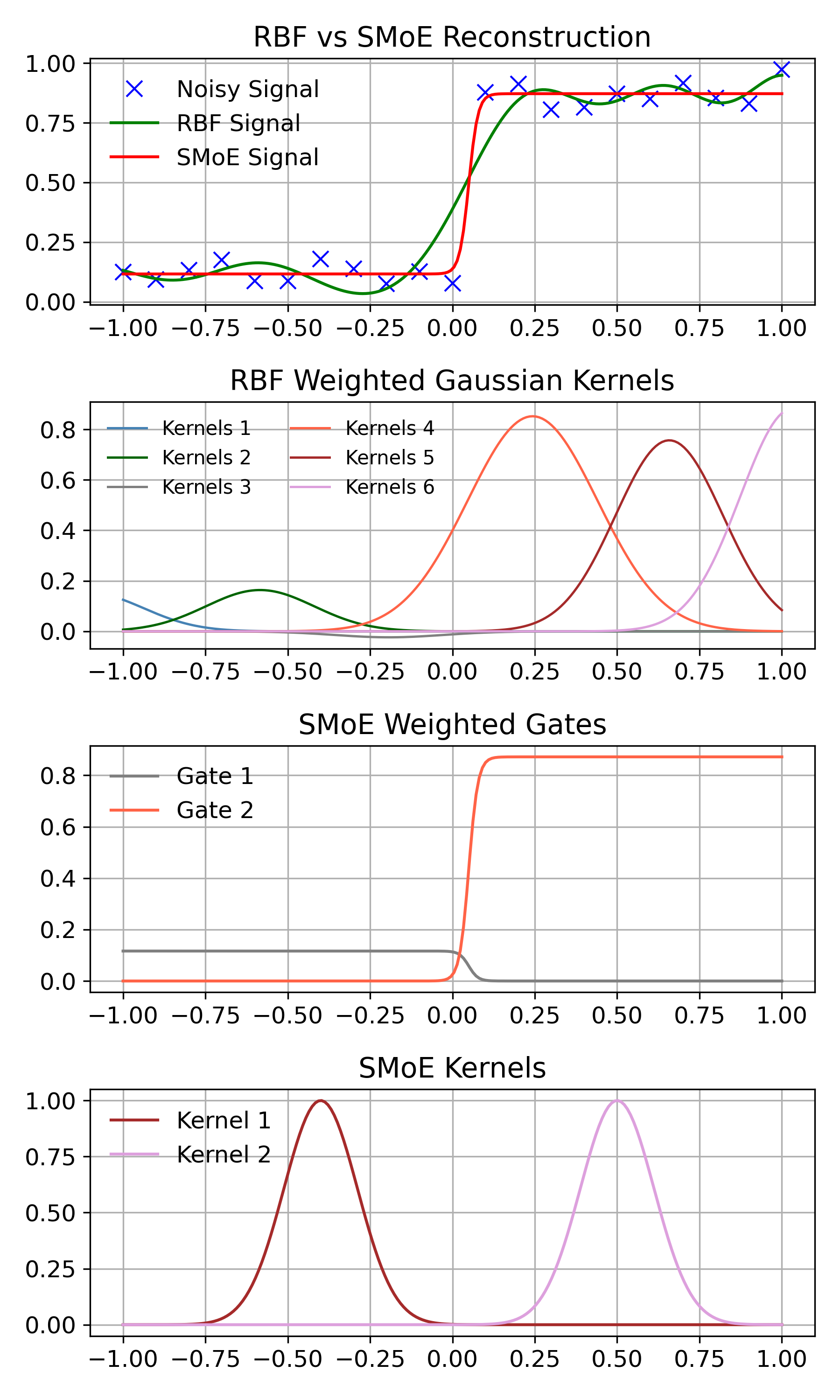}}
    \hfill
    \subfloat[Super-resolution with sharpening]{\includegraphics[width=0.49\linewidth]{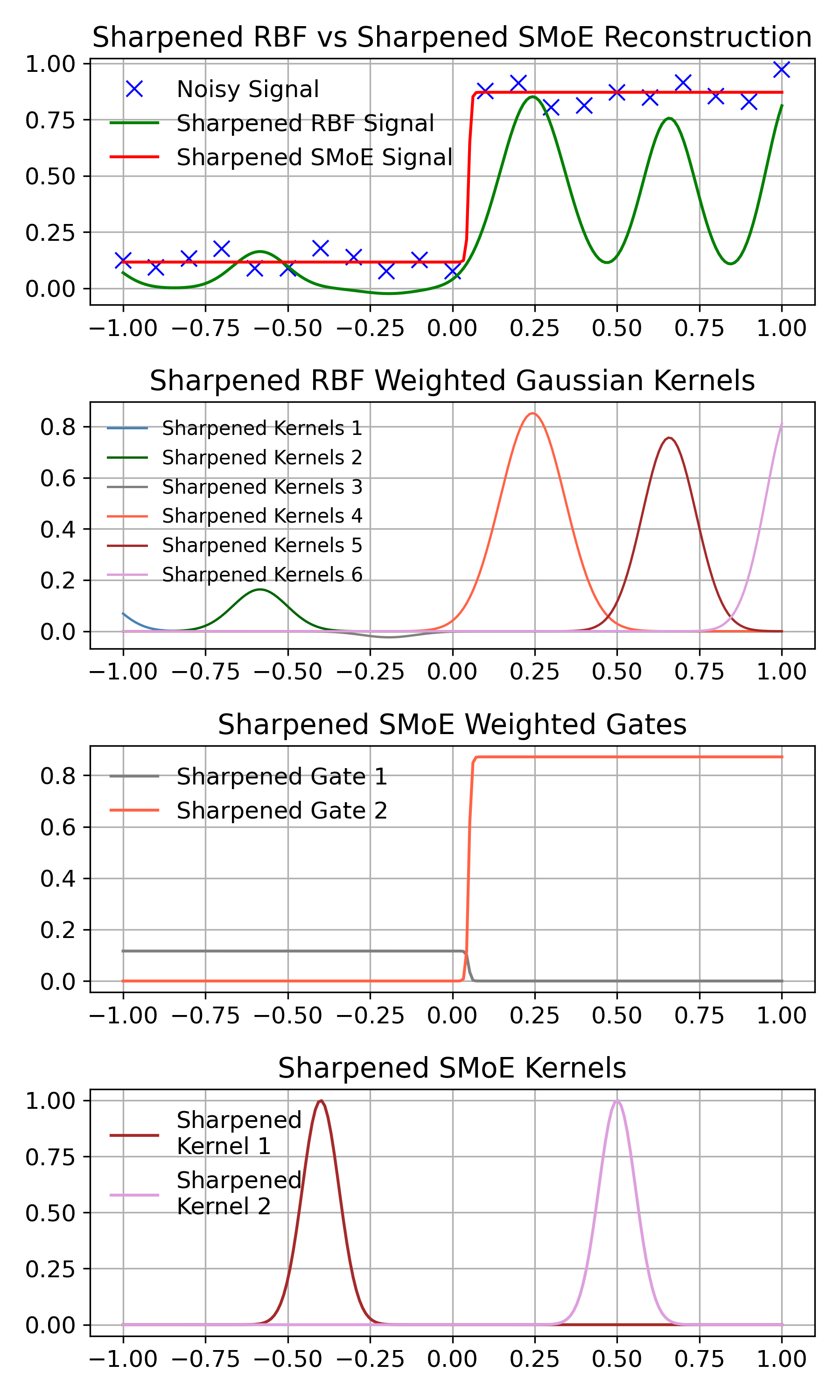}}
\vspace{-0.15cm}
\caption{Comparison of kernel behaviors and reconstruction results for the RBF and SMoE models \textcolor{black}{demonstrated on} denoising and super resolution \textcolor{black}{tasks}.}
\label{fig2}
\vspace{-0.4cm}
\end{figure}

\subsection{Different Capabilities of SMoE vs Gaussian Splatting RBF}
Fig. \ref{fig2} illustrates the difference in concepts and capabilities of RBF and SMoE frameworks on a noisy 1D signal in a block with 21 samples. As we shall subsequently see, the results of this simple 1D experiment carry over to Gaussian Splatting RBF and SMoE regression capabilities on complex 2D imagery. The findings readily explain the significant quality and computational gains of SMoE regression on clean and noisy images, as well as for super-resolution and sharpening tasks with kernel editing.

The \textcolor{black}{true} signal in Fig. \ref{fig2}(a) is a step function which models a sharp edge. Pixel values are between 0.1 and 0.9, thus flat with one sharp transition. Both locations and bandwidths of the Gaussian Splatting RBF and SMoE kernels were optimized using gradient descent. While two SMoE kernels effectively suffice to recover the sharp transition of the true signal from noisy data (27 dB PSNR), the RBF framework provides far inferior edge reconstruction even with six kernels (22 dB). In addition, the many kernels employed by the \textcolor{black}{\textit{dense}} RBF network attempt to model the noise in flat signal regions. This results in  \textcolor{black}{\textit{ringing}} artifacts. While more RBF kernels would provide better edge reconstruction, the capability of noise suppression would be further reduced. The SMoE model with its sparse representation recovers the flat signal efficiently, because two weighted soft-gating functions are employed for \textcolor{black}{the} reconstruction of the two flat regions. The location and bandwidths of the kernels have an indirect impact on the reconstruction. The gates provide ``global'' support with only two kernels. With RBF, the kernels contribute directly with ``local'' support. 

Fig. \ref{fig2}(b) depicts the super-resolution/sharpening capabilities of either method for the same true signal with noise.  Both RBF and SMoE provide a continuous regression function \(y_p(x)\). For super-resolution, pixel interpolation to any size, with regular or irregular pixel raster, can be easily produced by re-sampling \(y_p(x)\). With either method ``native'' sharpening of the resolution-enhanced signals can be performed by kernel editing -- reducing the bandwidths of the kernels by a sharpening factor. Fig. \ref{fig2}(b) illustrates how the new SMoE kernel bandwidths now result in excellent sharpened edge reconstruction, while preserving the flat regions perfectly. The reduced bandwidths of the RBF kernels result in slightly improved edge sharpening but drastically enhanced ringing. Some flat regions in the signal cannot be reproduced at all because kernel coverage between the kernels is significantly reduced. 

Based on the findings of this simple 1D experiment, we expect SMoE regression to outperform Gaussian Splatting RBF on 2D imagery, with
1) a sparser representation, 
2) with fewer artifacts, 3) faster training and reconstruction, 4) superior denoising capability, and 5) super-resolution and sharpening using kernel editing with fewer artifacts.

\section{Optimization for SMoE and GS Kernel Regression}
\subsection{Deep-Learning-Based Optimization }
Recently, several methods have been proposed to replace traditional gradient descent (GD) optimization with deep-learning approaches. \textcolor{black}{These methods train} neural networks to directly predict kernel parameters. For instance, Fleig et al. \cite{fleig_edge-aware_2022,fleig_steered_2023} trained an auto-encoder network to predict kernel parameters with a SMoE regressor according to Eq. (\ref{eq2}) as \textcolor{black}{the} decoder. This results in drastically faster optimization, replacing the cumbersome GD process. However, these approaches are \textcolor{black}{currently} applicable for a maximum of 16 kernels on small blocks of 8x8 or 16x16 pixels. 
Currently, \textcolor{black}{most} approaches use a fixed number of kernels for each block, which provides limited adaptation to content. Additionally, because kernel parameter prediction is based on small, isolated blocks, independent predictions across different blocks can introduce \textcolor{black}{blocking} effects, where discontinuities or artifacts may appear between blocks. 

\subsection{Global Optimization}
Global optimization \cite{bochinski_regularized_2018,jongebloed_sparse_2022} represents the forefront of optimization strategies for kernel regression and SMoE models. This method optimizes the model by considering the entire set of kernels simultaneously, rather than treating them individually or in isolated groups. The core idea is to ensure that all kernels in the model contribute to the reconstruction of each pixel, thereby capturing the holistic structure of the image. This strategy is also common in RBF type regression \cite{dale2007globalRBF}. Global optimization thus involves using all kernels to jointly reconstruct each pixel. Strategies exist to constrain the impact of kernels to the immediate neighborhood \cite{verhack2018progressive} to reduce the complexity of reconstruction. 

By involving all kernels in the optimization process, \textcolor{black}{global} optimization can achieve a high accuracy and coherence across the image. This holistic approach \textcolor{black}{preserves both global context and fine-grained} spatial relationships between pixels. 

The major drawback of \textcolor{black}{global} optimization is its computational intensity. Since every kernel must be considered for each pixel's reconstruction, the number of calculations increases dramatically. This results in \textcolor{black}{heavy computational demands}, making the process time-consuming and resource-intensive.

The need to update a large number of parameters jointly complicates the optimization process. This complexity can lead to slower convergence \textcolor{black}{and higher memory usage}, posing significant challenges for real-time applications and high-resolution image processing.

\subsection{Rasterized optimization}
Recently, rasterization \cite{rasterization_laine_2011,rasterization_schotz_2022} has become an increasingly valuable tool for enhancing gradient descent (GD) optimization in image processing tasks. One notable example is the GS for 2D images (GaussianImage \cite{zhang2024gaussianimage}), which leverages rasterization to boost the efficiency of GD optimization in \textcolor{black}{R}adial \textcolor{black}{B}asis \textcolor{black}{F}unction (RBF) \cite{ghosh_overview_2001} networks. 
Adopted from 3D GS \cite{kerbl_3d_2023}, this approach is similar to deep-learning-based optimization, \textcolor{black}{being} block-based in nature. Each image is divided into adjacent, non-overlapping blocks of 16x16 pixels. A very large number of 2D kernels \textcolor{black}{are} initialized over the image domain. A small subset of the kernels is identified as relevant for \textcolor{black}{regressing} a block prior to GD optimization, which allows fast and accurate GD optimization and block reconstruction. In this paper, we \textcolor{black}{adapt} the GS rasterized optimization strategy to SMoE regression, resulting in fast, sparse, and high-quality reconstruction.

\subsection{Hybrid EM-Based Parallel Rendering}

\textcolor{black}{In addition to gradient-descent-based rasterized optimization, prior work by Avramelos et al.~\cite{Avramelos2020highlyparallelsmoe} introduced a block-parallel rendering method for SMoE models, designed for real-time applications. Their method relies on a two-phase pipeline: a global Expectation-Maximization (EM) optimization followed by block-level parallel rendering. During optimization, Gaussian kernel parameters—position, orientation, and scale—are learned globally using EM. The rendering stage then employs a fixed-radius search around each pixel to determine which kernels to evaluate, enabling efficient per-pixel computation.}

\textcolor{black}{While their rendering is block-parallel, the training remains decoupled from this process and does not benefit from the same acceleration. In contrast, our approach rasterizes both rendering and optimization: kernel-to-block coverage is precomputed from the kernel's perspective, accounting for anisotropic scaling. This leads to substantial acceleration during both forward passes and backpropagation, fully integrating the benefits of rasterization into gradient-descent optimization. Unlike EM, our method offers flexible convergence and compatibility with standard deep learning pipelines.}

\begin{figure*}
    \centering
    \subfloat[Pipeline of Rasterized SMoE]{\includegraphics[width=0.7\linewidth]{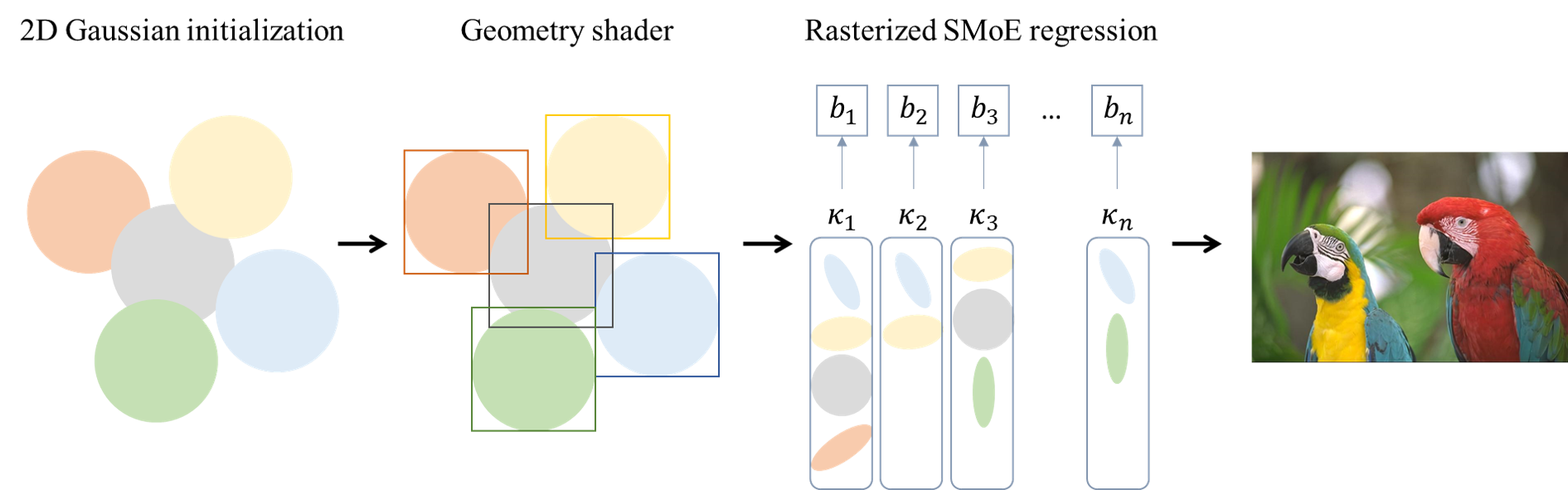}}
    \hfill
    \centering
    \subfloat[Geometry shader]{\includegraphics[width=0.3\linewidth]{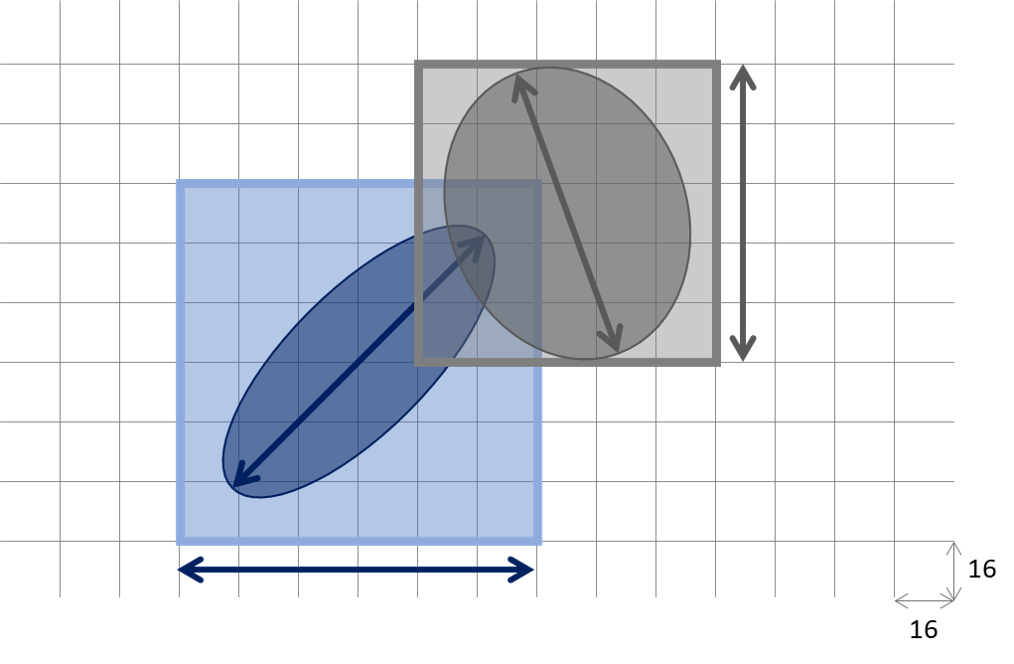}}
\caption{(a) \textcolor{black}{Left:} 2D Gaussian initialization with five colored circles representing Gaussian kernels. \textcolor{black}{Middle:} The bounding boxes indicate the coverage of blocks affected by the corresponding Gaussian kernels. \textcolor{black}{Right:} The block $b_n$ is reconstructed by the corresponding affected kernel set $\mathcal{K}_n$. Kernels are represented by distinct colors and shapes. (b) Gaussian kernels are represented as ellipses with varying axes. The coverage of affected blocks is shown by square boxes aligned with the centers of the kernels (ellipses), where the box side length matches the long axis of the corresponding kernel (ellipse).}
\label{fig3}
\vspace{-0.4cm}
\end{figure*}

\section{Rasterized SMoE}
Rasterized SMoE (R-SMoE) regression \textcolor{black}{limits} the set of kernels involved in the reconstruction of each pixel in a $16\times16$ pixel block $b_i$ to a subset \( \mathcal{K}_i \) of the entire set $\mathcal{K}$, where \( i \) corresponds to the block index and $\mathcal{K}_i \subseteq \mathcal{K}$. 
This localization is achieved by truncating the Gaussian kernels’ spatial support—similar to the approach in Gaussian Splatting—thereby making each kernel effectively local rather than global. Unlike the original SMoE formulation where kernels have global influence, this truncation is essential to enable efficient block-wise kernel selection and rasterized processing.

The pipeline for the proposed rasterized SMoE is illustrated in Fig. \ref{fig3} (a). This pipeline \textcolor{black}{mirrors that of} \cite{zhang2024gaussianimage}, except that the RBF regression is replaced with the SMoE gating framework (c.f. Equ. (\ref{eq3}) and (\ref{eq4})).

\subsection{Gaussian kernel initialization}
In the Gaussian kernel initialization stage (in Fig. \ref{fig3} (a), left), a set  $\mathcal{K}$  of round 2D kernels is randomly distributed across the image. The bandwidth of each kernel is determined by  $\frac{W}{L}$, where $W$ represents the image width and $L$ represents the total number of kernels.

\subsection{Geometry shader}

For each 16x16 pixel block, each Gaussian kernel of the set  $\mathcal{K}$ is given a confidence ellipse corresponding to a 99\% confidence interval of its 2D Gaussian distribution. This ellipse approximates the region in which the kernel has significant influence, with its major and minor axes derived from the kernel’s covariance matrix. For the mathematical intricacies—including axis derivation via eigen-decomposition—see \cite{genz2009computation}.

Fig. \ref{fig3} (b) shows the process of the geometry shader. We derive the bounding box for each Gaussian kernel, with sides equal to the major axis of the confidence ellipse. This bounding box ensures that all pixels within it are highly influenced by the corresponding Gaussian kernel. Although some pixels outside the confidence ellipse may still fall within the bounding box, these pixels are discarded during gradient calculations to maintain computational efficiency. Note that \textcolor{black}{prior to} GD optimization, the kernels are \textcolor{black}{initialized} as round \textcolor{black}{in the Geometry Shader stage} (Fig. \ref{fig3} (a)).
\subsection{Subset Kernel Selection for Each Block}

For each Gaussian kernel $K_i$, the bounding box is used to identify the blocks that intersect with it. Each intersected block $b_n$ is recorded as being affected by this Gaussian kernel $K_i$. Consequently, for each block, we compile a subset of kernels $\mathcal{K}_n$ that are deemed to have a significant impact on its reconstruction. To reflect this in the regression function, Eq. (3) is revised as follows:
\begin{equation}
y_p(x) = \sum_{j \in \mathcal{K}_n} m_j(x) \cdot w_j(x).
\label{eq5}
\end{equation}
In this equation, $\mathcal{K}_n$ represents the subset of kernels that significantly affect block $b_n$. Fig. \ref{fig3} (a) depicts the different steered kernels selected as relevant for each block $b_n$ during GD. It is important to note that $\mathcal{K}_n$ is not independent of the subsets $\mathcal{K}_{n'}$ for neighboring blocks $b_{n'}$. This interdependence arises because the bounding boxes of different Gaussian kernels may overlap, leading to shared kernels between adjacent blocks. Consequently, the parameters of these shared kernels are optimized according to the adjacent blocks, rather than just in a single block.

This overlap and interdependence between subsets $\mathcal{K}_n$ and $\mathcal{K}_{n'}$ ensure that the spatial correlations between pixels are preserved, addressing \textcolor{black}{a key} limitation of conventional block-based methods that \textcolor{black}{typically fail to preserve} global context. By focusing only on the most influential kernels for each block, our method reduces the computational burden while maintaining the integrity of spatial relationships across the image.
  
\section{Rasterized SMoE for Denoising}

While previous SMoE methods have employed block-based representations for denoising, either through gradient descent (GD) \cite{ozkan_steered_2023} or deep learning (DL) \cite{fleig_steered_2023} optimization, these approaches are limited by their block constraints. R-SMoE promises a more efficient and reliable representation, combining the advantages of global representation with the flexibility of local processing. R-SMoE does not confine the influence of kernels to a small block, allowing each kernel to contribute beyond its immediate neighborhood.

One significant challenge when applying GS for 2D images and SMoE models to denoising tasks is that, when applied to smaller blocks of 16x16 pixels, these models attempt to accurately reconstruct not only the signal but also part of the noise. A promising strategy is then to employ overlapping GS \textcolor{black}{blocks} or R-SMoE blocks and reconstruct the pixels using multiple models in a multi-model approach \cite{ozkan_steered_2023}. This yields a natural extension: multi-model inference in R-SMoE (MM-RSMoE), which we explore for denoising.

\subsection{Theoretical Considerations - Denoising a Single Block}

To establish the theoretical foundation for denoising using R-SMoE, we start by considering a single block. 

Suppose an image block pixel \( y(x) \) is represented by a SMoE model with \( L \) kernels, and is corrupted by an additive noise signal \( \epsilon(x) \), assumed to come from a covariance-stationary, zero-mean noise process \( \{\mathcal{E}(\mathbf{x})\} \). The observed noisy image is given by \( y_r(\mathbf{x}) = y(\mathbf{x}) + \epsilon(\mathbf{x}) \), with zero mean \( \mu_\epsilon(\mathbf{x}) = 0 \) and noise variance  \( \delta_\epsilon^2(\mathbf{x}) = \delta_\epsilon^2 \). The SMoE model can be expressed as:
\begin{equation}
    \hat{y}(\mathbf{x}) = \sum_{j=1}^L \left( m_j(\mathbf{x}) + \epsilon_j(\mathbf{x}) \right) \cdot w_j(\mathbf{x}),
\label{eq6}
\end{equation}
where the noise process \( \{\mathcal{E}(\mathbf{x})\} \) is assumed to be statistically independent of the original signal process \( \{Y(\mathbf{x})\} \). In this model, the parameter \( m_j \) is estimated in region \( R_j \) as:
\begin{equation}
    \hat{m}_j = \hat{\mu}_{Y_j} =\sum_{r=1}^{ R_j} \hat{y}(\mathbf{x}_r) \cdot  \frac{w_j(\mathbf{x}_r)}{\sum_{k=1}^{ R_j} w_j(\mathbf{x}_k)},
\label{eq7}
\end{equation}
which leads to the estimation  \( \hat{m}_j = m_j + \hat{\mu}_{\epsilon_j} \). The uncertainty in the estimation is measured by the variance:
\begin{equation}
    \delta_{\hat{m}_j}^2 = \frac{\delta_\epsilon^2}{M_j},
\label{eq8}
\end{equation}
where \( M_j = \sum_{k=1}^{ R_j} w_j(\mathbf{x}_k) \) and \( \hat{\mu}_{\epsilon_j} = \sum_{r=1}^{ R_j} \epsilon(\mathbf{x}_r)\cdot\frac{w_j(\mathbf{x}_r)}{M_j} \) represents the number of samples covered by the gating function \( w_j \) and the bias for estimating the parameter $m_j$, respectively. 

Thus, with a sufficient number of noise samples \( M_j \) captured by a gating function \( w_j(\mathbf{x}) \), we can expect the estimate \( \hat{m}_j \) to approach the true value \( m_j \) without bias. The variance \( \delta_{\hat{m}_j}^2 \) measures the uncertainty in this estimate, which diminishes as \( M_j \) increases. \textcolor{black}{In short}, the larger the number of noise samples \( M_j \) covered by a gating function \( w_j \), the less biased the estimate of \( m_j \), leading to a more accurate model inference. Inspect the simple example in Fig. 2(a) to understand how the two gates each capture noise samples.

\subsection{Segmentation with modified DBSCAN for Enhanced Denoising}
Recall that each Gaussian kernel \( K_j \) influences a set of blocks \( B_j \), determined by its bounding box. The total number of pixels \( M_j \) used for denoising estimation scales with the number of affected blocks, \( M_j = |B_j| \cdot b \), where \( b \) is the pixels per block. Increasing \( |B_j| \) thus directly expands the data supporting each kernel’s denoising.

\( |B_j| \) is limited by local neighborhoods unless we increase the spatial extent of the kernel’s influence. Here, segmentation provides a powerful strategy: by grouping pixels into coherent regions \( R_j \), we effectively enlarge the spatial domain associated with each kernel.
Specifically, kernels are assigned within segments \( R_j \), each containing a set of pixels larger than a single block. Because kernels are distributed within these larger segments, the number of blocks each kernel affects, \( |B_j| \), grows approximately proportional to the segment size \( |R_j| \), where \( |R_j| \) denotes the number of pixels in segment \( R_j \)\textcolor{black}{,} divided by the number of kernels per segment \( n_k \):

\begin{equation}
|B_j| \approx \frac{|R_j|}{n_k},
\label{eq9}
\end{equation}
where \( n_k = \frac{L}{N} \), with \( L \) total kernels and \( N \) segments.
Therefore, by using segmentation to increase \( |R_j| \), we effectively increase \( |B_j| \), allowing each kernel to leverage a larger, more informative pixel set during denoising. This targeted expansion preserves spatial coherence while improving noise robustness and reconstruction fidelity.

We employ \textcolor{black}{modified DBSCAN} \cite{li_segmentation-based_2023}, a region-based clustering method, to generate these meaningful segments from pixel RGB similarity. More details on this segmentation-based initialization can be found in the cited work.

\subsection{Multi-Model Fusion}

In practical scenarios, each SMoE model may introduce additional noise, referred to as model noise \( \mathbf{e}_j(\mathbf{x}) \). This noise arises because the SMoE model is trained independently on each block of data, with the variance \( \delta_{\mathbf{e}_j}^2 \) being unknown and depending on factors like random initialization. As a result, the noisy image $y'$ is more accurately modeled:
\begin{equation}
    y'(\mathbf{x}) = \hat{y}(\mathbf{x}) + \mathbf{e}(\mathbf{x}) ,
\label{eq10}
\end{equation}
where \( \mathbf{e}(\mathbf{x}) \) represents model noise\textcolor{black}{s},  \( \hat{y}(\mathbf{x}) \) is the prediction at pixel location \( x \), and \( \mu_{e} = 0\) with \(\delta_{e}^2 \) remains unknown.
Consider a multi-model approach\textcolor{black}{—multiple overlapping blocks,} each \textcolor{black}{providing} a prediction for a particular noisy pixel. Given $H$ SMoE models, the multi-model image $y_m$ is expressed by fusing the outputs of $H$ SMoE models, here by averaging individual predictions:
\begin{equation}
    y_m(\mathbf{x}) = \frac{1}{H}\sum_{h=1}^H  y'_h(\mathbf{x}) =\frac{1}{H}\sum_{h=1}^H  \hat{y}_{h}(\mathbf{x}) + \mathbf{e}_h(\mathbf{x}) ,
\label{eq11}
\end{equation}
where  \( \hat{y}_h(\mathbf{x}) \) is the prediction from the \( h \)-th model at pixel location \( x \), \(\mu_{e_h} = \mu_{e} = 0\) and \(\delta_{e_h}^2 = \delta_{e}^2 \) for \( h = 1, \dots, H \) are assumed to be unknown.
We assume that \textcolor{black}{the prediction} \( \hat{y}_{h}(\mathbf{x}) \) is nearly constant across different \( h \), we can approximate \(\hat{y}_{1}(\mathbf{x}) \approx \hat{y}_{2}(\mathbf{x}) \approx \dots \approx \hat{y}_{H}(\mathbf{x}) \approx \hat{y}(\mathbf{x})\), and thus,
\begin{equation}
    y_m(\mathbf{x}) \approx \hat{y}(\mathbf{x}) + \frac{1}{H} \sum_{h=1}^H \mathbf{e}_h(\mathbf{x}).
\label{eq12}
\end{equation}
Given the formula for $\hat{y}(\mathbf{x})$ in Eq. (\ref{eq6}), the estimated parameter \( \hat{m}_j \) \textcolor{black}{is updated. This update now includes a term that accounts for model noise: }  
\begin{equation}
    \hat{m}_j = m_j + \hat{\mu}_{\epsilon_j} + \hat{\mu}_{e_j},
    \label{eq13}
\end{equation}
where \( \hat{\mu}_{e_j} = \frac{1}{H}\sum_{h=1}^H \mu_{\mathbf{e}_h}=\mu_{\mathbf{e}}=0\). The corresponding variance in the estimate is updated to:
\begin{equation}
    \delta_{\hat{m}_j}^2 = \frac{\delta_\epsilon^2}{M_j} + \text{Var}\left(\frac{1}{H} \sum_{h=1}^H \mathbf{e}_h(\mathbf{x}) \right) = \frac{\delta_\epsilon^2}{M_j} + \frac{\delta_e^2}{H}.
    \label{eq14}
\end{equation}
By using multiple SMoE models, each trained on different noisy signal blocks that have similar underlying noisy signals \( y(\mathbf{x}) + \epsilon(\mathbf{x}) \), we can effectively reduce the impact of model noise. As \( H \) approaches infinity, the contribution of model noise diminishes, leading to a more accurate and unbiased estimation of the parameters \( m_j \). Thus, multi-model fusion \textcolor{black}{not only} mitigates model noise \textcolor{black}{but also} enhances the \textcolor{black}{SMoE framework’s denoising capacity}.

\textcolor{black}{We propose a multi-model approach that uses block-overlapping SMoE models.} For this purpose, a block window is shifted over the image horizontally and vertically with shift-displacements in the range [2, 4, 6, \dots, 16] pixels. For the pixels in each window a SMoE regression model is trained. This results in [64, 32, 16, \dots, 1] multi-hypotheses (predictions) generated for each original image pixel.

\section{Experimental Setting}

In this section, we describe the datasets, evaluation metrics, implementation details, and baselines used to assess the performance of the proposed Rasterized SMoE \textcolor{black}{(R-SMoE)} model.

\subsection{Datasets, Metrics, and Baselines}
Our experiments focus on two core tasks: image regression and denoising. For both tasks, we use the widely adopted Kodak dataset \cite{kodak}, which comprises 24 high-quality color images at a resolution of 768×512 pixels. This dataset serves as a standard benchmark for evaluating image fidelity and perceptual quality. 
\textcolor{black}{To further assess the robustness of our method, we additionally include the DIV2K dataset \cite{DIV2K}, consisting of 100 high-resolution images (1060×768) with diverse natural content. This dataset allows us to examine performance under more varied textures and structures.}

To quantify performance, we employ three commonly used image quality metrics. Peak Signal-to-Noise Ratio (PSNR) and Structural Similarity Index Measure (SSIM) assess pixel-wise accuracy and structural preservation, respectively. Additionally, we report Learned Perceptual Image Patch Similarity (LPIPS), which provides a perceptual measure of visual similarity based on deep feature representations, offering complementary insight beyond PSNR and SSIM.

We benchmark the R-SMoE model against several state-of-the-art baselines. For image regression, \textcolor{black}{we compare against} GS for 2D images (GaussianImage) \cite{zhang2024gaussianimage}, \textcolor{black}{a recent rasterized RBF method based on Gaussian Splatting} (referred to as GS-RBF); the global SMoE (GSMoE) model \cite{verhack_steered_2020,bochinski_regularized_2018}; and the Radial Basis Function (RBF) approach. While the Radial Basis Function (RBF) framework encompasses any function of the form \( \phi(\mathbf{x}) = \hat{\phi}(\|\mathbf{x} - \mathbf{u}\|) \), we follow common practice and adopt the Gaussian RBF, defined as \( \phi(r) = \exp(-\gamma r^2) \), consistent with the implementation in \cite{ghosh_overview_2001}. For denoising, we use the well-established BM3D algorithm \cite{bm3d_2007} as the primary baseline.

\begin{table*}[t]
\caption{\textcolor{black}{\textbf{Quantitative comparison of R-SMoE, G-SMoE, GaussianImage, and RBF for two kernel settings (2000 and 10000). Metrics include PSNR, SSIM, LPIPS, encoding/decoding time, FPS, FLOPs, and GPU memory usage. R-SMoE consistently achieves higher quality with faster runtime and lower computational cost.}}}
\centering
\setlength{\tabcolsep}{3pt}
\begin{tabular}{p{.13\linewidth}wc{.07\linewidth}wc{.07\linewidth}wc{.07\linewidth}wc{.07\linewidth}wc{.12\linewidth}wc{.12\linewidth}wc{.07\linewidth}wc{.07\linewidth}wc{.07\linewidth}}
\hline
\multicolumn{10}{c}{\textbf{Total number of available kernels: 2000}} \\
\hline
Method & Avg. kernel$\downarrow$ & PSNR(dB)$\uparrow$ & SSIM$\uparrow$ & LPIPS$\downarrow$ & Encode time$\downarrow$ & Decode time$\downarrow$ & FPS$\uparrow$ & FLOPs$\downarrow$ & GPU usage$\downarrow$\\
\hline
RBF\cite{ghosh_overview_2001} & 2000 & 27.20 & 0.7279 & 0.4156 & 1746s & 0.349s & 2.86 & 4000 & \textbf{876 MB} \\
GSMoE\cite{verhack_steered_2020} & 2000 & \textbf{28.10} & \textbf{0.7662} & \textbf{0.3628} & \textbf{1725s} & \textbf{0.346s} & \textbf{2.89} & 4000 & 878 MB \\
\hline
GaussianImage\cite{zhang2024gaussianimage} & 56 & 27.59 & 0.7422 & 0.3983 & 98s & 2.2ms & 449 & 112 & \textbf{768 MB} \\
R-SMoE & \textbf{40} & \textbf{27.99} & \textbf{0.7618} & \textbf{0.3635} & \textbf{69s} & \textbf{1.8ms} & \textbf{528} & \textbf{80} & 770 MB\\
\hline
\multicolumn{10}{c}{\textbf{Total number of available kernels: 10000}} \\
\hline
Method & Avg. kernel$\downarrow$ & PSNR(dB)$\uparrow$ & SSIM$\uparrow$ & LPIPS$\downarrow$ & Encode time$\downarrow$ & Decode time$\downarrow$ & FPS$\uparrow$ & FLOPs$\downarrow$ & GPU usage$\downarrow$\\
\hline
RBF\cite{ghosh_overview_2001} & 10000 & 32.38 & 0.8926 & 0.2065 & \textbf{7486s} & \textbf{1.51s} & \textbf{0.66} & 20000 & \textbf{1290 MB} \\
GSMoE\cite{verhack_steered_2020} & 10000 & \textbf{32.98} & \textbf{0.9040} & \textbf{0.1900} & 7673s & 1.53s & 0.65 & 20000 & 1292 MB \\
\hline
GaussianImage\cite{zhang2024gaussianimage} & 69 & 32.82 & 0.8997 & 0.1954 & 104s & 2.4ms & 415 & 140 & \textbf{778 MB} \\
R-SMoE & \textbf{51} & \textbf{33.13} & \textbf{0.9074} & \textbf{0.1769} & \textbf{81s} & \textbf{2.2ms} & \textbf{443} & \textbf{102} & 780 MB \\
\hline
\end{tabular}
\label{table1}
\end{table*}

\begin{table*}

\caption{\textcolor{black}{\textbf{Quantitative comparison on the DIV2K dataset across different total kernel pool sizes. Metrics include PSNR, SSIM, and LPIPS, showing how increasing the kernel pool improves reconstruction quality. R-SMoE achieves competitive quality with significantly fewer kernels.}}}
\centering
\setlength{\tabcolsep}{3pt}
\begin{tabular}{p{.13\linewidth}wc{.07\linewidth}wc{.07\linewidth}wc{.07\linewidth}wc{.07\linewidth}wc{.12\linewidth}wc{.12\linewidth}wc{.07\linewidth}wc{.07\linewidth}wc{.07\linewidth}}
\hline
Method & Avg. Kernel$\downarrow$ & PSNR$\uparrow$ & SSIM$\uparrow$ & LPIPS$\downarrow$ & Encode time$\downarrow$ & Decode time$\downarrow$ & FPS$\uparrow$ & FLOPs$\downarrow$& GPU usage$\downarrow$\\
\hline
\multicolumn{10}{c}{{Total Available Kernels @ 2000}}\\
\hline
GaussianImage \cite{zhang2024gaussianimage} & 13 & \textbf{28.05} & 0.84 & 0.30 & 156s & 1.69ms & 591&26&\textbf{768} MB\\
R-SMoE & \textbf{10} & 27.77 & 0.84 & 0.30 & \textbf{116s} & \textbf{0.64ms} & \textbf{1564}&\textbf{20}&770 MB\\
\hline
\multicolumn{10}{c}{{Total Available Kernels @ 10000}}\\
\hline
GaussianImage \cite{zhang2024gaussianimage}& 73 & \textbf{36.60} & 0.96 & 0.10 & 165s & 2.09ms & 479 &146&\textbf{778} MB\\
R-SMoE & \textbf{57} & 36.49 & 0.96 & 0.10 & \textbf{136s} & \textbf{1.13ms} & \textbf{882} &\textbf{114}&780 MB\\
\hline

\end{tabular}
\label{table2}
\end{table*}


\begin{figure}[t]
    \centering
    \setcounter{subfigure}{0}
    \includegraphics[trim={0.cm 0.cm 0.cm 0cm},clip,width=1\linewidth]{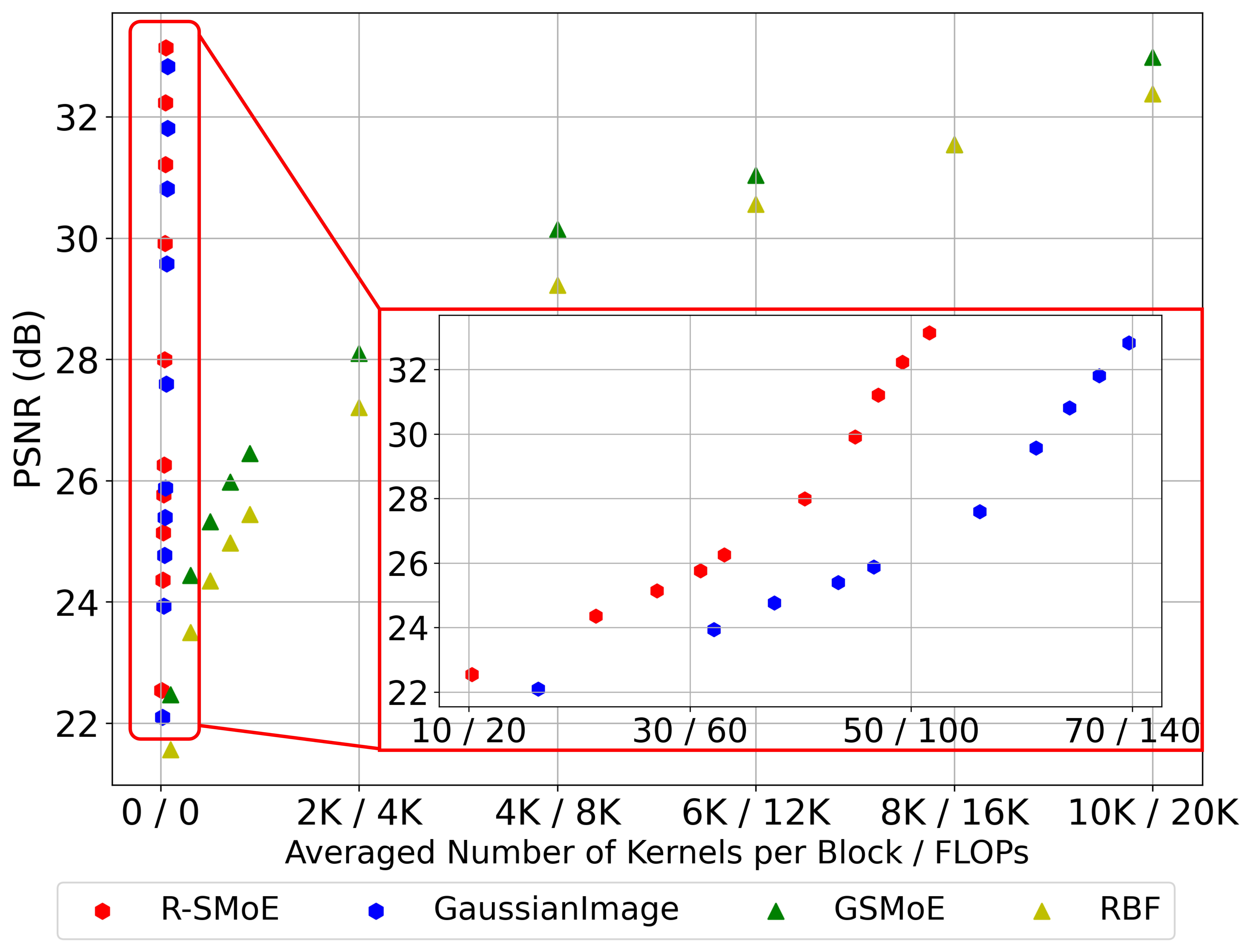}
    \vspace{-1em}
    \caption{PSNR versus average number of kernels per block. The x-axis shows the average number of kernels used to render each block, highlighting the efficiency of selecting a subset rather than employing all available kernels. The figure offers a comprehensive comparison across four models: R-SMoE, G-SMoE, GaussianImage, and RBF, demonstrating the substantial reduction in computational demand achieved by R-SMoE. A \textcolor{black}{red rectangle} emphasizes a detailed comparison between R-SMoE and GaussianImage, illustrating subtle yet crucial differences in FLOPs. \textcolor{black}{Our method delivers up to a 6 dB PSNR improvement over state-of-the-art methods when operating under similar average kernel counts, underscoring both quality and efficiency gains.}}
    \vspace{-0em}
    \label{fig4}
\end{figure}

\begin{figure*}[h]
\centering
\includegraphics[trim={0.1cm 0.2cm 0.1cm 0cm},clip,width=1\linewidth]{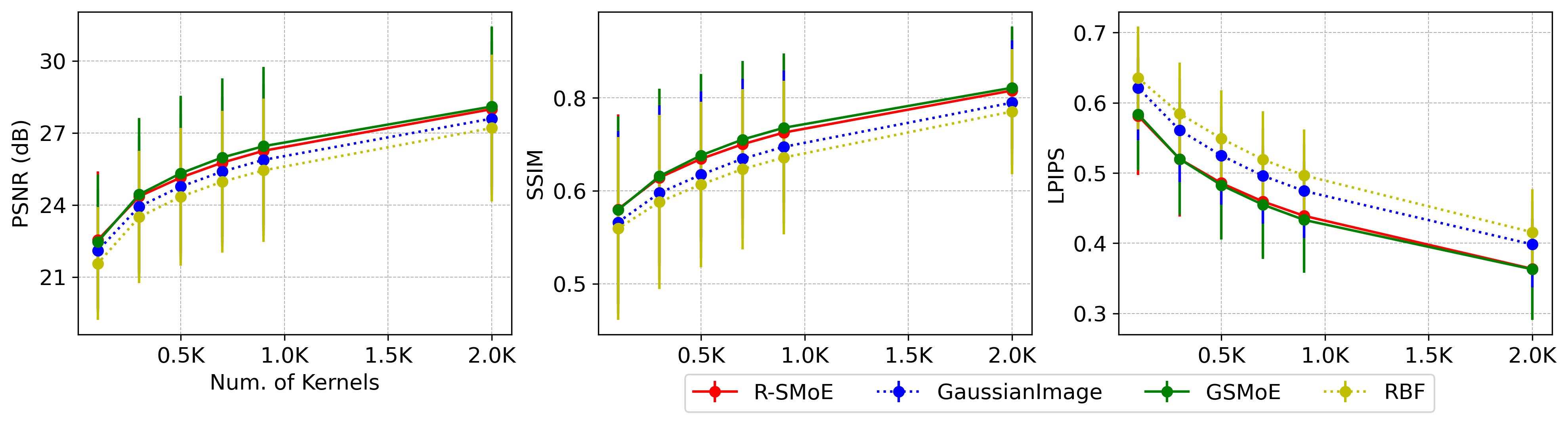}
\vspace{-1em}
\caption{Performance scaling \textcolor{black}{with kernel pool size on PSNR, SSIM, and LPIPS. The curves show how image quality improves with larger kernel pools while highlighting R-SMoE’s ability to maintain high visual fidelity with fewer kernels, demonstrating its efficiency–quality trade-off advantage.}}
\vspace{-0em}
\label{fig5}
\end{figure*}

\subsection{Implementation Details}

The R-SMoE model is implemented within the GS-Splat framework \cite{zhang2024gaussianimage}, extended with specialized CUDA kernels to perform rasterization through a weighted sum of gating functions. The covariance of the 2D Gaussians is parameterized using Cholesky factorization.

\textcolor{black}{\textbf{Data:}} \textcolor{black}{All images are processed in RGB space at their original resolution. The Kodak dataset consists of 24 images at 768$\times$512 pixels, while the DIV2K dataset contains high-quality images with resolutions around 1060$\times$768 pixels.}

\textcolor{black}{\textbf{Kernel Initialization:}} \textcolor{black}{Each regression starts with an initial pool of $L$ kernels, randomly initialized with a fixed scaling factor of 5 pixels. The gating mechanism dynamically determines the average number of active kernels per block. For denoising tasks, segmentation boundaries are determined using modified DBSCAN \cite{li_segmentation-based_2023}, with pixel difference thresholds of 10 and 20.}

\textcolor{black}{\textbf{Training Protocol:}} All models are optimized using Adam \cite{diederik2015adam} for 10,000 iterations. The learning rates are set to 0.01 for Gaussian centers ($\mu$), 0.001 for covariance matrices ($\Sigma$), and 0.001 for the expert outputs ($m$). \textcolor{black}{The learning rate for $\mu$ decays exponentially to 0.00001. Early stopping is not applied; convergence is empirically determined after 10,000 iterations.}

\textcolor{black}{\textbf{Hardware and Runtime:}} \textcolor{black}{All experiments are conducted on NVIDIA A4000 GPUs (16 GB) running Ubuntu 20.04 with CUDA 11.8 and PyTorch 2.1. FLOPs are reported following the same per-pixel measurement methodology as \cite{zhang2024gaussianimage}.}

\begin{figure*}[h!]
    \centering
    \begin{minipage}{0.2\linewidth}  
        \includegraphics[width=\linewidth]{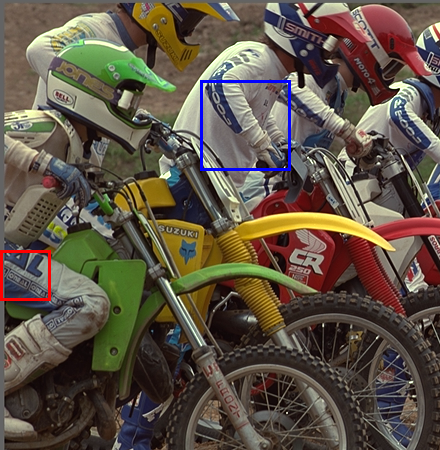}  
    \end{minipage}%
    \hfill
    \begin{minipage}{0.8\linewidth}  
        \includegraphics[width=\linewidth]{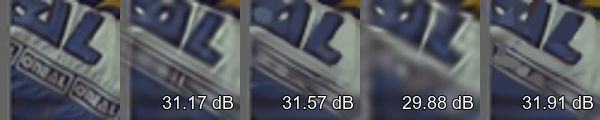}
        \includegraphics[width=\linewidth]{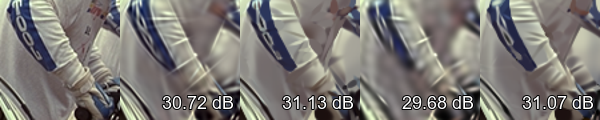}
    \end{minipage}
        \begin{minipage}{0.2\linewidth}  
        \includegraphics[width=\linewidth]{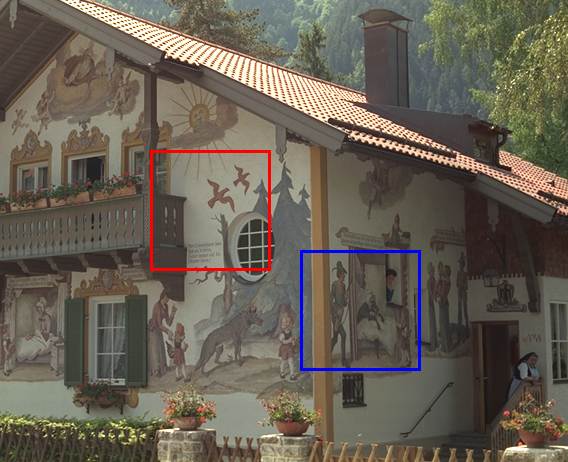}  
    \end{minipage}%
    \hfill
    \begin{minipage}{0.8\linewidth}  
        \includegraphics[width=\linewidth]{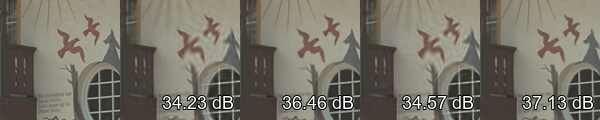}
        \includegraphics[width=\linewidth]{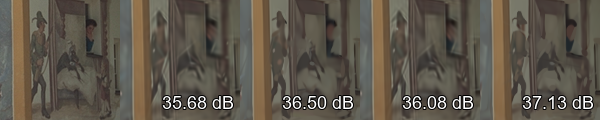}
    \end{minipage}
    \\
    \begin{minipage}{0.2\linewidth}  
        \includegraphics[width=\linewidth]{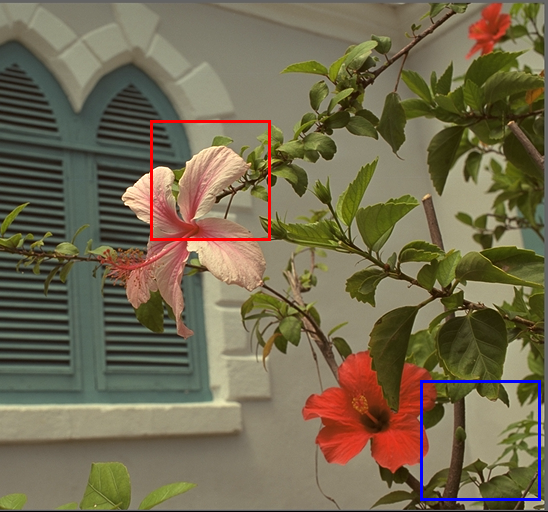}  
    \end{minipage}%
    \hfill
    \begin{minipage}{0.8\linewidth}  
        \includegraphics[width=\linewidth]{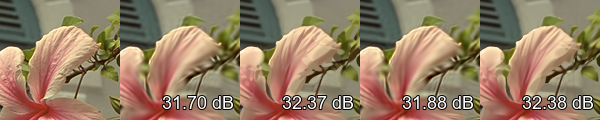}
        \includegraphics[width=\linewidth]{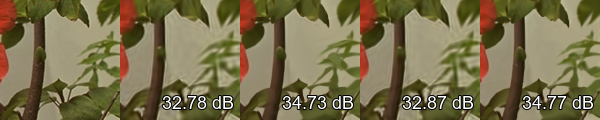}
    \end{minipage}
    \\
    \begin{tabular}{>{\centering}p{0.18\linewidth}>{\centering}p{0.135\linewidth}>{\centering}p{0.135\linewidth}>{\centering}p{0.135\linewidth}>{\centering}p{0.135\linewidth}>{\centering}p{0.135\linewidth}}
         & Original & RBF & GSMoE & GaussianImage & R-SMoE
    \end{tabular}

    \caption{Visualization of image regression results \textcolor{black}{at 6000 kernels for} RBF \cite{ghosh_overview_2001}, GSMoE \cite{verhack_steered_2020}, GaussianImage \cite{zhang2024gaussianimage}, and the proposed R-SMoE. \textcolor{black}{Two regions from the original image (left) are cropped and enlarged to highlight differences in edge sharpness and fine-structure reconstruction.}}
    \label{fig6}
\end{figure*}

\begin{figure}[!t]
    \centering
    \includegraphics[width=0.32\linewidth]{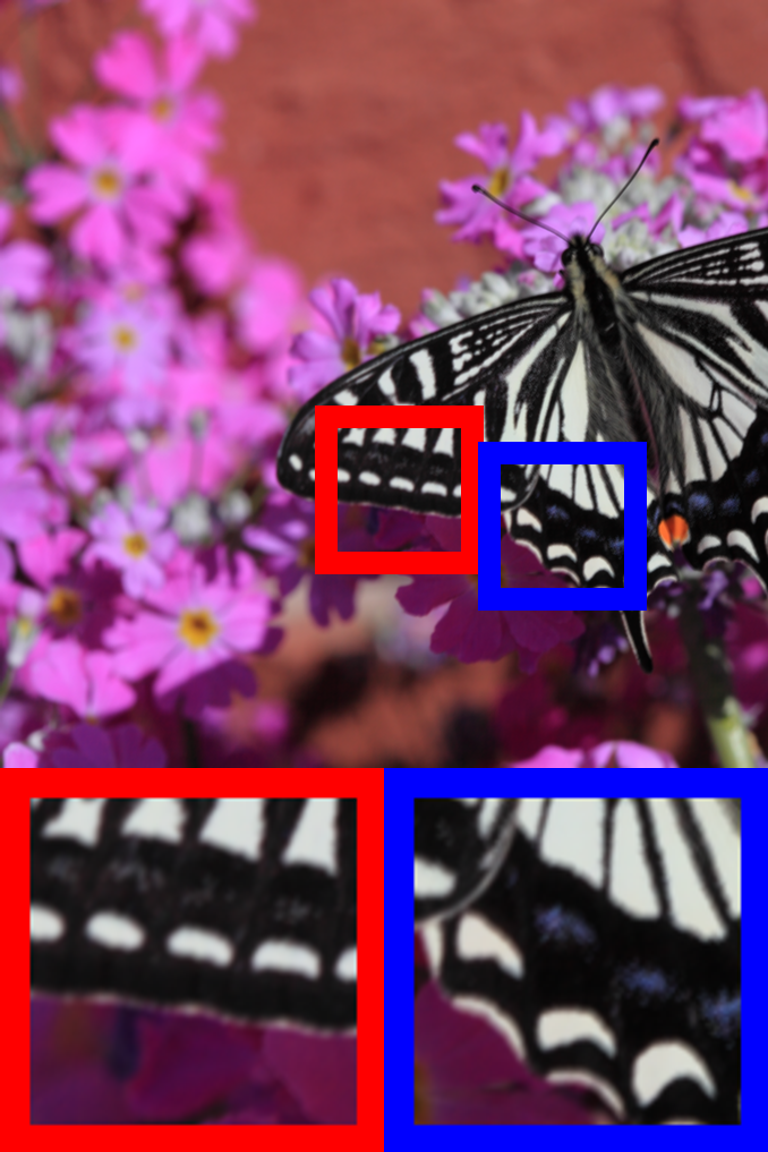}
    \hfil
    \includegraphics[width=0.32\linewidth]{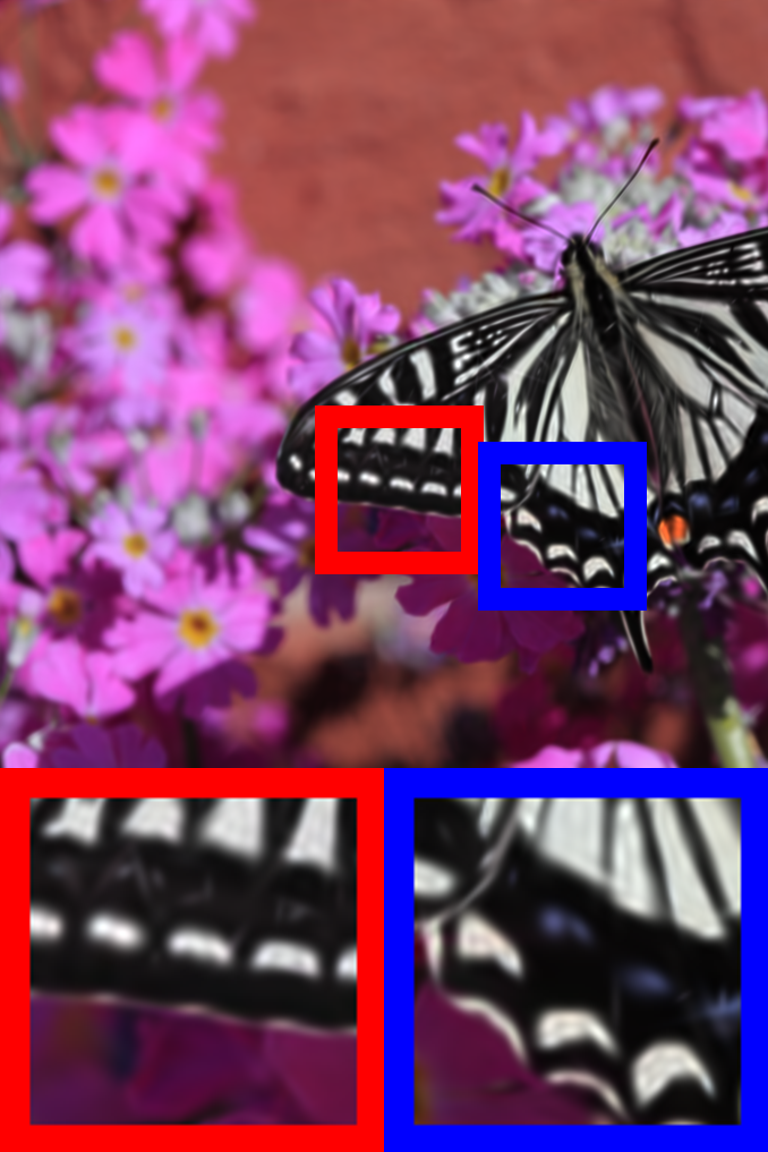}
    \hfil
    \includegraphics[width=0.32\linewidth]{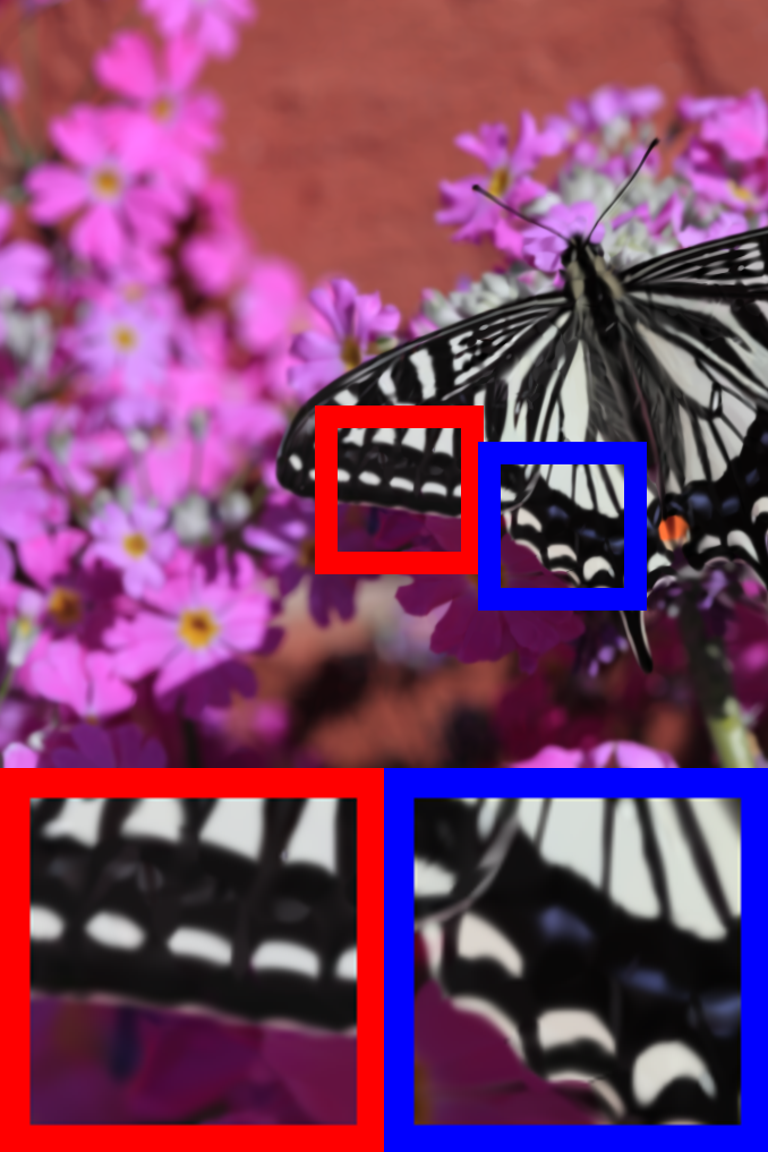}
    \\
    \subfloat[Ground Truth]{\includegraphics[width=0.32\linewidth]{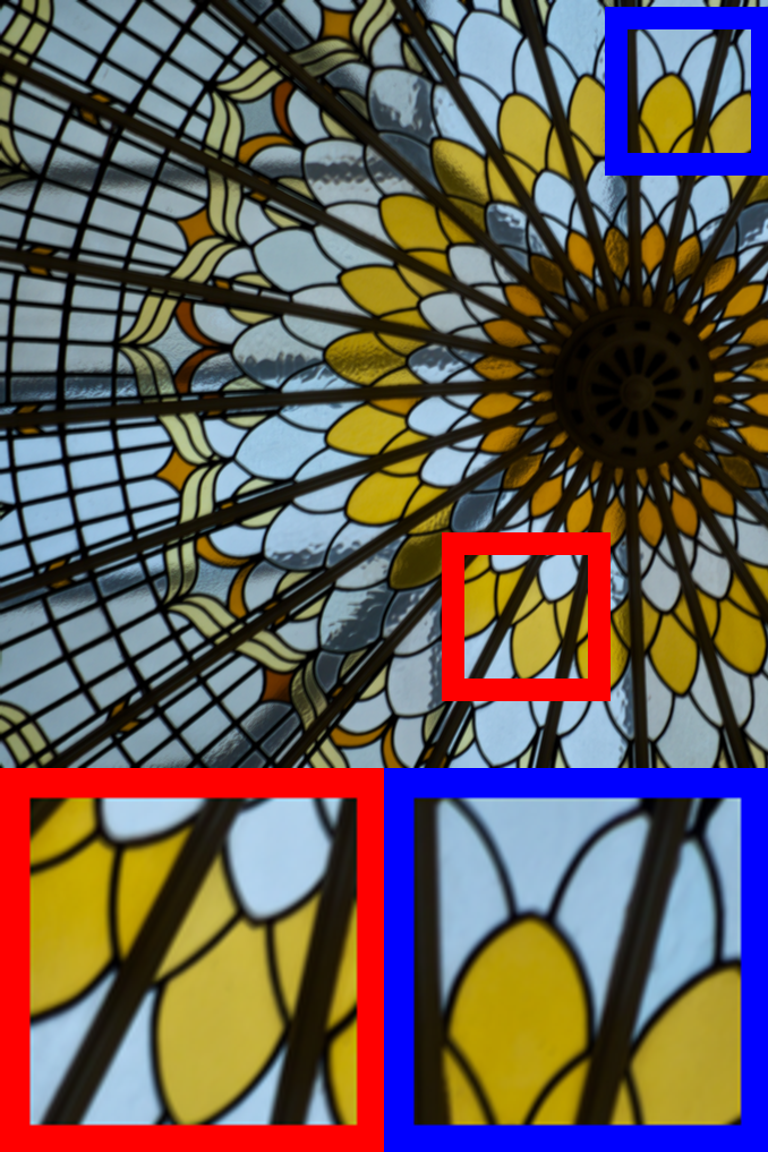}}
    \hfil
    \subfloat[GaussianImage]{\includegraphics[width=0.32\linewidth]{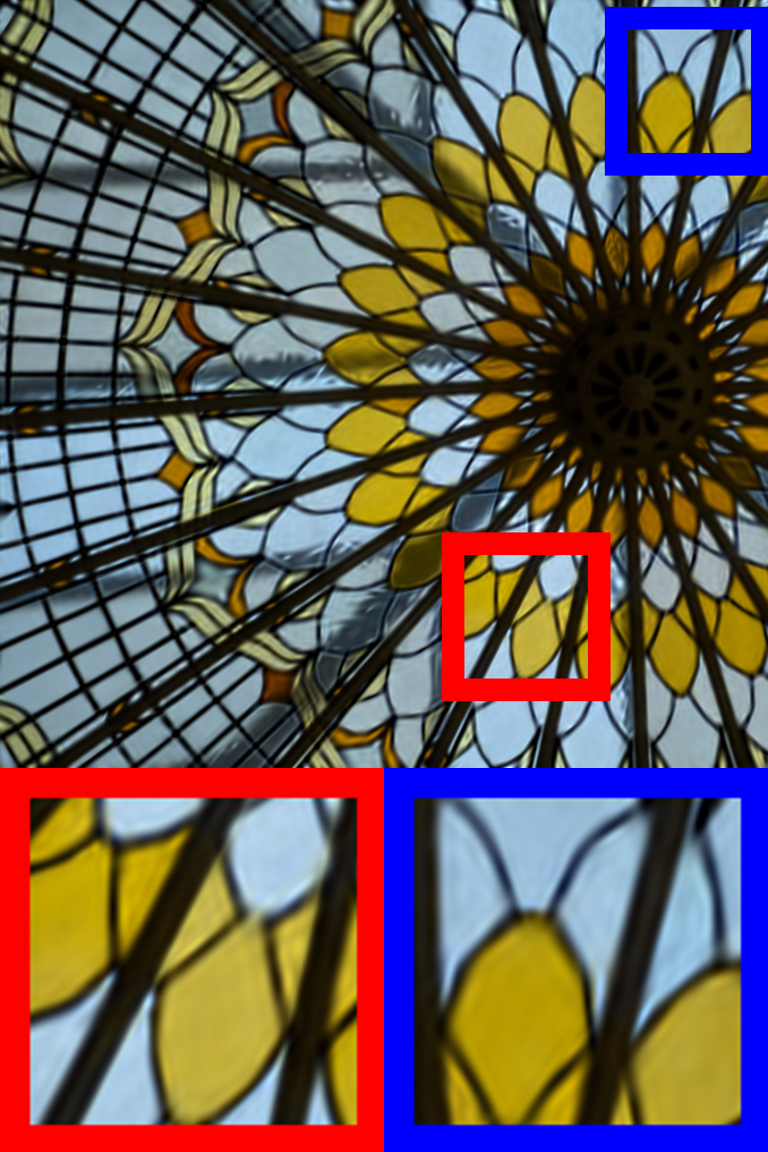}}
    \hfil
    \subfloat[R-SMoE]{\includegraphics[width=0.32\linewidth]{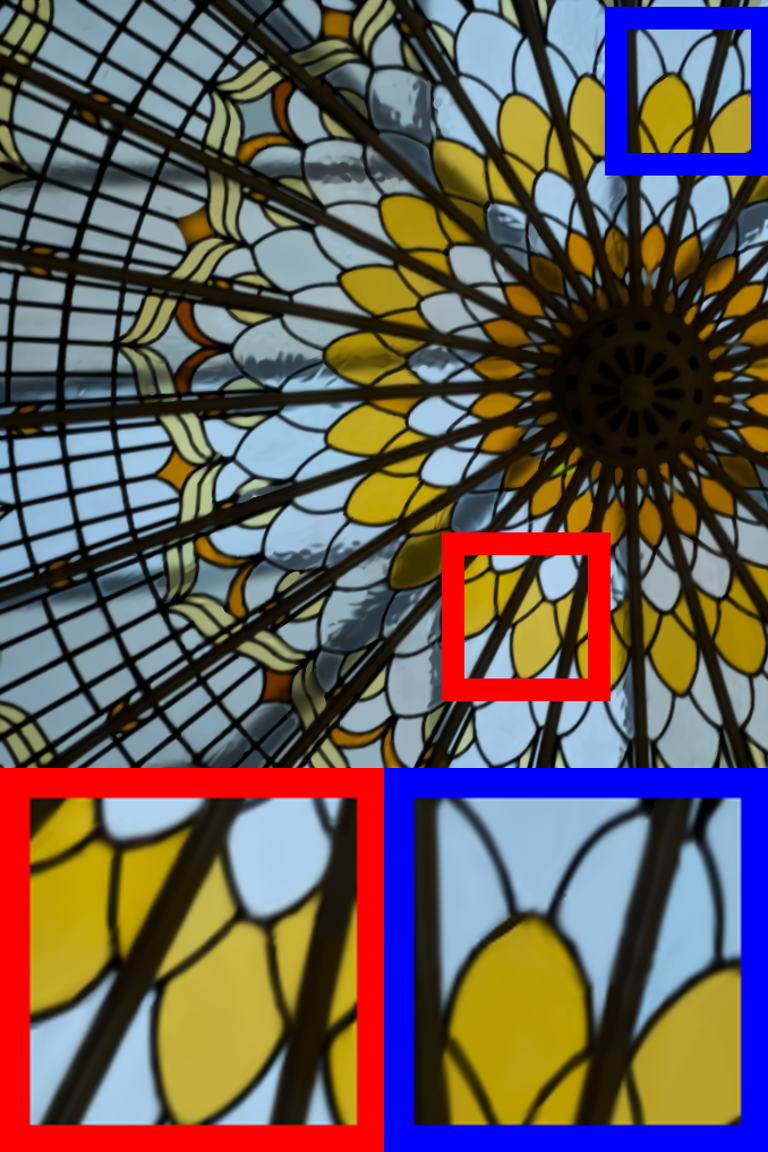}}
    \caption{\textcolor{black}{\textbf{Visual comparison on the DIV2K dataset.} R-SMoE achieves comparable perceptual quality to GaussianImage \cite{zhang2024gaussianimage} while producing sharper edges and smoother textures with fewer kernels. Error maps are scaled for visibility.}}
    \label{fig7}
\end{figure}
\section{Results}

\subsection{Image Regression}

Table \ref{table1} and Fig. \ref{fig4} \textcolor{black}{reveal} the performance \textcolor{black}{gains of the} rasterized SMoE (R-SMoE) model \textcolor{black}{against} GaussianImage \cite{zhang2024gaussianimage}, GSMoE \cite{verhack_steered_2020, bochinski_regularized_2018}, and RBF \cite{ghosh_overview_2001}. Table \ref{table1} \textcolor{black}{summarizes} reconstruction quality metrics, while Fig. \ref{fig4} demonstrates \textcolor{black}{reductions in} computational complexity and memory usage. Bold values \textcolor{black}{highlight} the \textcolor{black}{best method per group}: global methods (above the line) and rasterized methods (below).

\textcolor{black}{Global SMoE} (GSMoE) significantly outperforms global Gaussian Splatting (RBF) regression by all quality measures. This accounts for a 0.6–0.9 dB gain on average, depending on the number of Gaussians used. Rasterized SMoE (R-SMoE), on the other hand, \textcolor{black}{matches GSMoE's quality} while improving the encoding and decoding times as well as GPU memory load by orders of magnitude. Reconstruction of images/frames per second improved from about 3 FPS to around 530 FPS or 0.7 FPS to 443 FPS, depending on the number of available Gaussians.

Compared to previously published work on Rasterized 2D Gaussian Splatting (GaussianImage), R-SMoE improves quality by 0.3–0.4 dB and significantly improves encoding and decoding times. 
\textcolor{black}{This trend is consistent across the additional DIV2K dataset, where R-SMoE achieved nearly identical SSIM and LPIPS values to GaussianImage (differences below 0.01) while maintaining competitive PSNR (only 0.3 dB lower on average). Crucially, R-SMoE preserved its computational advantage, yielding 17\% faster training and 45\% higher rendering FPS for high-bpp and 25\% faster training with 62\% higher rendering FPS for low-bpp. These results confirm that the proposed method generalizes well to diverse, high-resolution content.}

The  \textcolor{black}{\textit{Avg. Kernels}} column in Table \ref{table1} represents the average number of kernels per block, \textcolor{black}{contrasting with} the total number of available kernels in the kernel pool. Although the total number of available kernels is fixed at 2000 and 10000 in the experiments, the \textcolor{black}{\textit{Avg. Kernels}} indicate the average subset actively utilized for processing each block. Correspondingly, the table also reports floating-point operations (FLOPs) per pixel, which quantify the actual processing load. Because fewer active kernels reduce FLOPs, these two metrics jointly illustrate how kernel sparsity translates into computational efficiency, enabling faster decoding and lower resource consumption. 

Fig. \ref{fig4} \textcolor{black}{extends this with} rate-distortion \textcolor{black}{curves plotting} PSNR against the average number of kernels per block and FLOPs. It provides a broad comparison of four models (R-SMoE, G-SMoE, GaussianImage, and RBF). An inset \textcolor{black}{zooms in on} R-SMoE \textcolor{black}{versus} GaussianImage. Unlike Table \ref{table1}, a \textcolor{black}{fixed-kernel} snapshot, Fig. \ref{fig4} highlights the efficiency of our method in terms of both computational load and resource utilization by reporting the average number of selected kernels per block. 

GaussianImage also uses rasterized optimization to cut down training time relative to global methods. However, it requires significantly more training time than R-SMoE due to its higher average kernels per block. This is because GaussianImage does not employ gating and introduces less sparsity compared to the proposed R-SMoE, which utilizes a gating network, \textcolor{black}{cf. }Section II.

Fig. \ref{fig4} \textcolor{black}{exposes} two primary advantages. First, our method uses fewer active kernels and cuts computational load compared to GaussianImage. This efficiency translates into faster decoding, as confirmed by Table \ref{table1}, enabling more efficient processing without sacrificing performance. Second, our method requires \textcolor{black}{fewer} GPU resources, allowing for a more lightweight implementation that still achieves exceptional results. Under comparable averaged kernel counts and FLOPs per pixel — and consequently, comparable per-pixel processing complexity — our approach achieves a PSNR improvement of up to 6 dB relative to GaussianImage at approximately 55 average kernels per block. This gain underscores the efficacy of our approach, offering both computational efficiency and superior performance.

We further demonstrate the performance consistency and the trade-off between PSNR and the total number of available kernels in Fig. \ref{fig5}. As the total number of available kernels increases, the performance improves for two main reasons. First, a larger kernel pool allows for a larger subset of kernels to be selected for each block, enabling finer control over the reconstruction process. Second, with a larger pool, different blocks can select distinct, non-overlapping subsets of kernels. This reduces the amount of ``shared'' kernel usage across blocks, meaning each kernel is used to update a smaller number of blocks. As a result, each kernel's updates are more focused, leading to more precise and specialized kernel adjustments, which ultimately \textcolor{black}{enhance} both quality and computational efficiency.

While global SMoEs tend to achieve slightly higher PSNR under limited kernel counts—for example, a 0.1 dB gain at 2000 kernels as shown in Table \ref{table1}—their training and rendering times increase substantially. In contrast, R-SMoE exploits rasterization during optimization, \textcolor{black}{reducing} both training duration and rendering time while delivering \textcolor{black}{comparable} quantitative performance \textcolor{black}{as} shown in Table \ref{table1}. Crucially, despite GaussianImage emphasizing its capability to achieve over 1000 FPS in its original paper, our R-SMoE implementation outperforms it in rendering speed under a fair comparison on \textcolor{black}{the} same hardware. Although FPS values \textcolor{black}{vary} depending on \textcolor{black}{the} GPU model \textcolor{black}{used}, our results consistently demonstrate faster rendering, highlighting the efficiency of our approach. This advantage stems from R-SMoE’s design: unlike global methods such as GSMoE and RBF, whose computational and memory costs scale linearly with the total number of kernels $L$, R-SMoE restricts computation to a relevant subset of kernels via rasterized lookup, ensuring scalability and speed. \textcolor{black}{Its speed comes from limiting computation to a small, spatially relevant subset of kernels per block}—making training and decoding time dependent primarily on local kernel density, not total model size. This enables efficient parallelization and consistent runtime performance.

In addition to computational complexity, Table \ref{table1} presents a comparison of GPU memory usage, highlighting the significant advantage of the rasterized approaches. Since rasterized methods process \textcolor{black}{only parts of} kernels rather than \textcolor{black}{full} kernels per block, the memory required for rendering is reduced by approximately 28\% compared to global approaches. This substantial reduction in memory usage further underscores the efficiency of R-SMoE for rendering tasks.


Fig. \ref{fig6} presents reconstructed images sampled from the test set, showing significant enhancement in visual quality with R-SMoE. R-SMoE effectively preserves high-frequency details, such as edges with \textcolor{black}{fewer} Gaussians compared to GaussianImage. GaussianImage and RBF, being non-gating kernel methods, produce expected artifacts, such as ringing boundary effects of Gaussian kernels. These “needle-like” distortions arise from long-bandwidth Gaussians bleeding across regions, as illustrated in Fig. \ref{fig2}. In contrast, R-SMoE’s gating-driven structure effectively suppresses such artifacts, yielding cleaner, more coherent reconstructions. The same edge-aware reconstruction property without ringing artifacts is \textcolor{black}{also observed in} the GSMoE results. \textcolor{black}{A similar trend appears in the DIV2K dataset (Fig. \ref{fig7}), where R-SMoE achieves comparable perceptual quality to GaussianImage while maintaining sharper edges and smoother textures with significantly fewer active kernels. Error maps confirm that both methods yield similar residual distributions, validating the robustness of the proposed approach.}

\begin{figure}[t]
\centering
\subfloat[2000 Kernels]{\includegraphics[trim={0.25cm 0.cm 0.cm 0.cm},clip,width=0.505\linewidth]{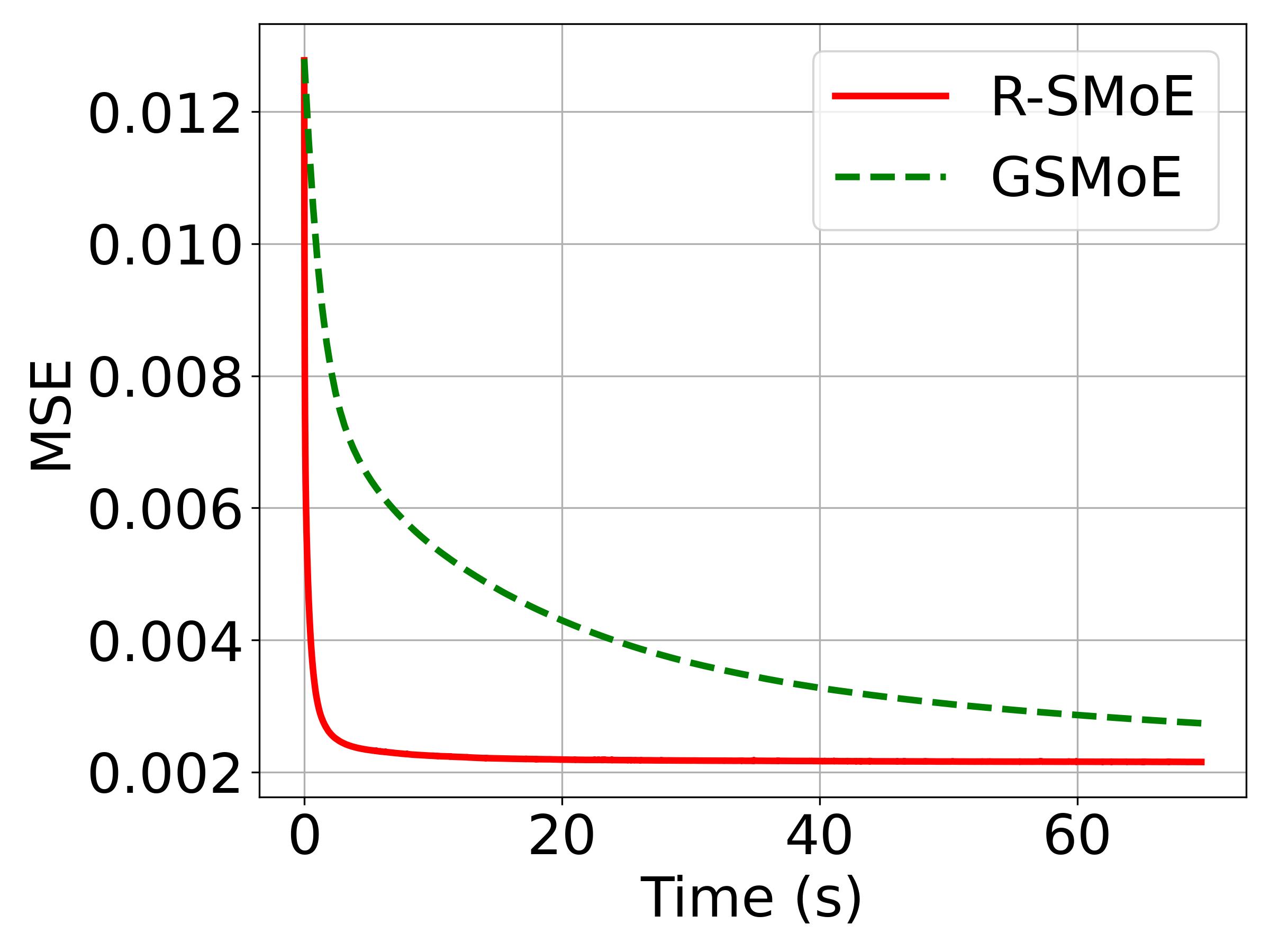}}
\subfloat[10000 Kernels]{\includegraphics[trim={1.18cm 0.cm 0.cm 0.cm},clip,width=0.48\linewidth]{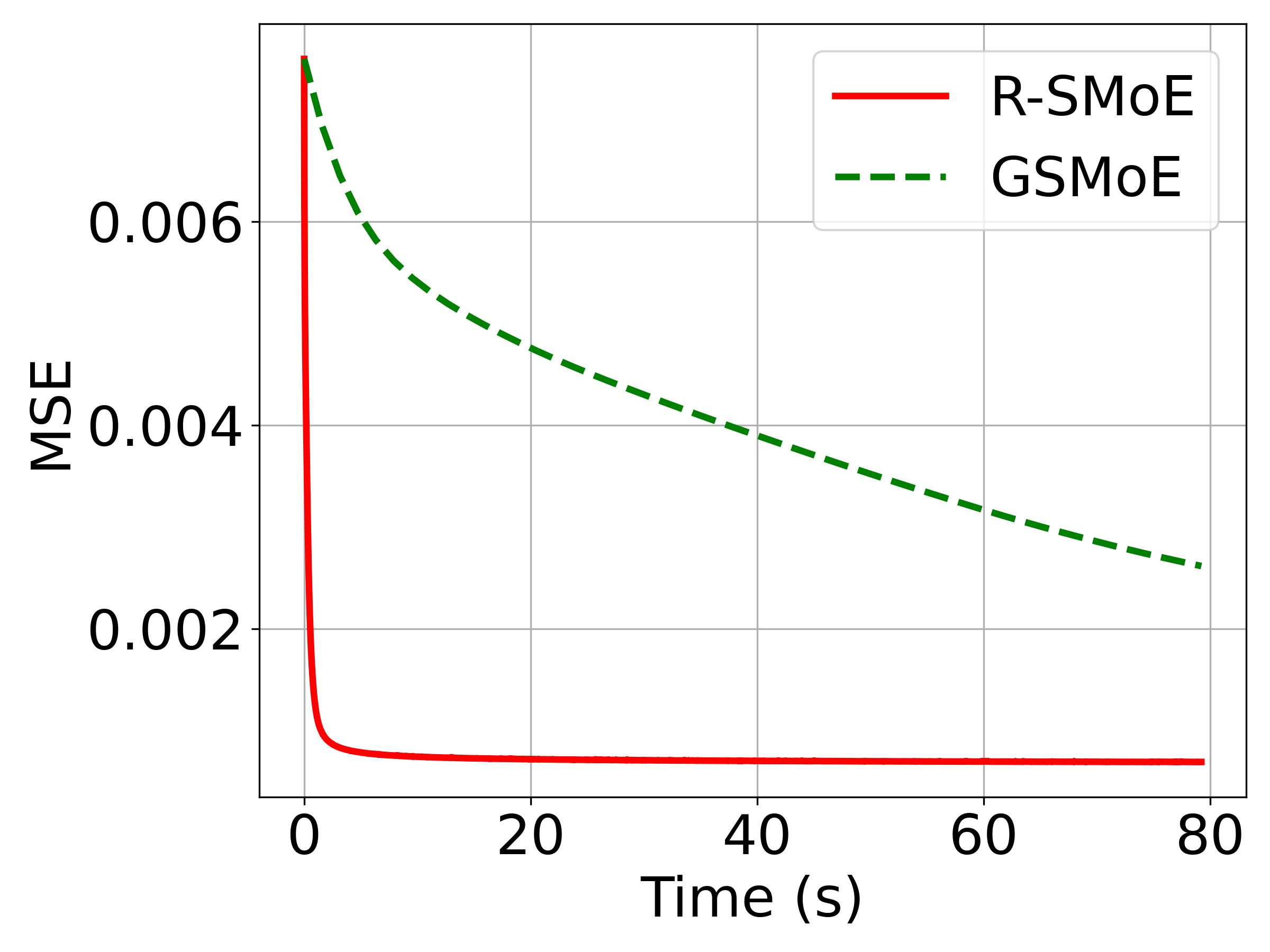}}
\vspace{0em}
\caption{Loss convergence of R-SMoE and GSMoE \textcolor{black}{models over training epochs, illustrating accelerated convergence of R-SMoE due to rasterization.}}
\vspace{-0em}
\label{fig8}
\end{figure}
\begin{figure}[t]
\centering
\subfloat[Four models]{\includegraphics[trim={0cm 0.cm 0.0cm 0cm},clip,width=0.50\linewidth]{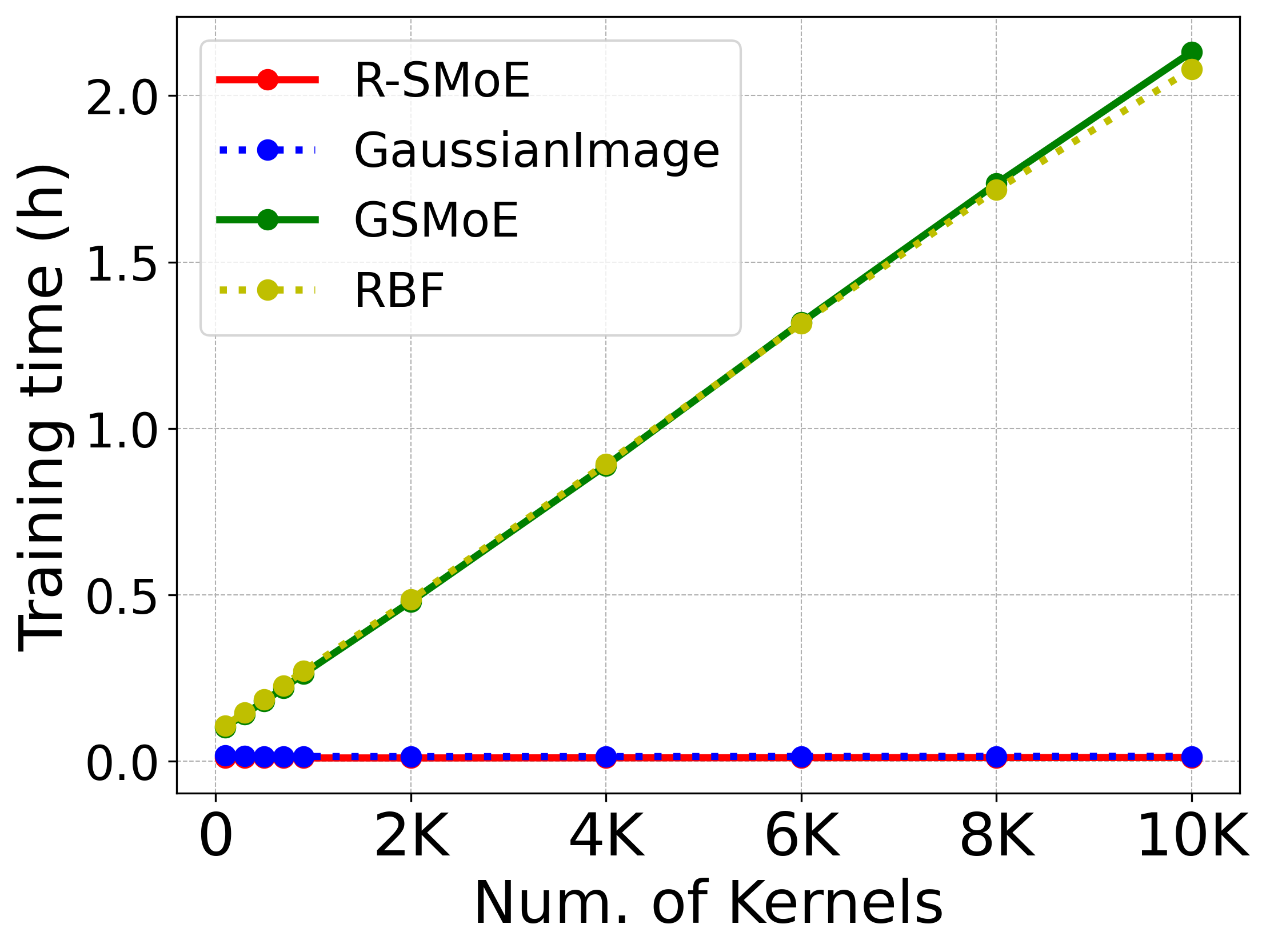}}
\subfloat[R-SMoE and GaussianImage]{\includegraphics[trim={0.0cm 0.cm 0.0cm 0cm},clip,width=0.50\linewidth]{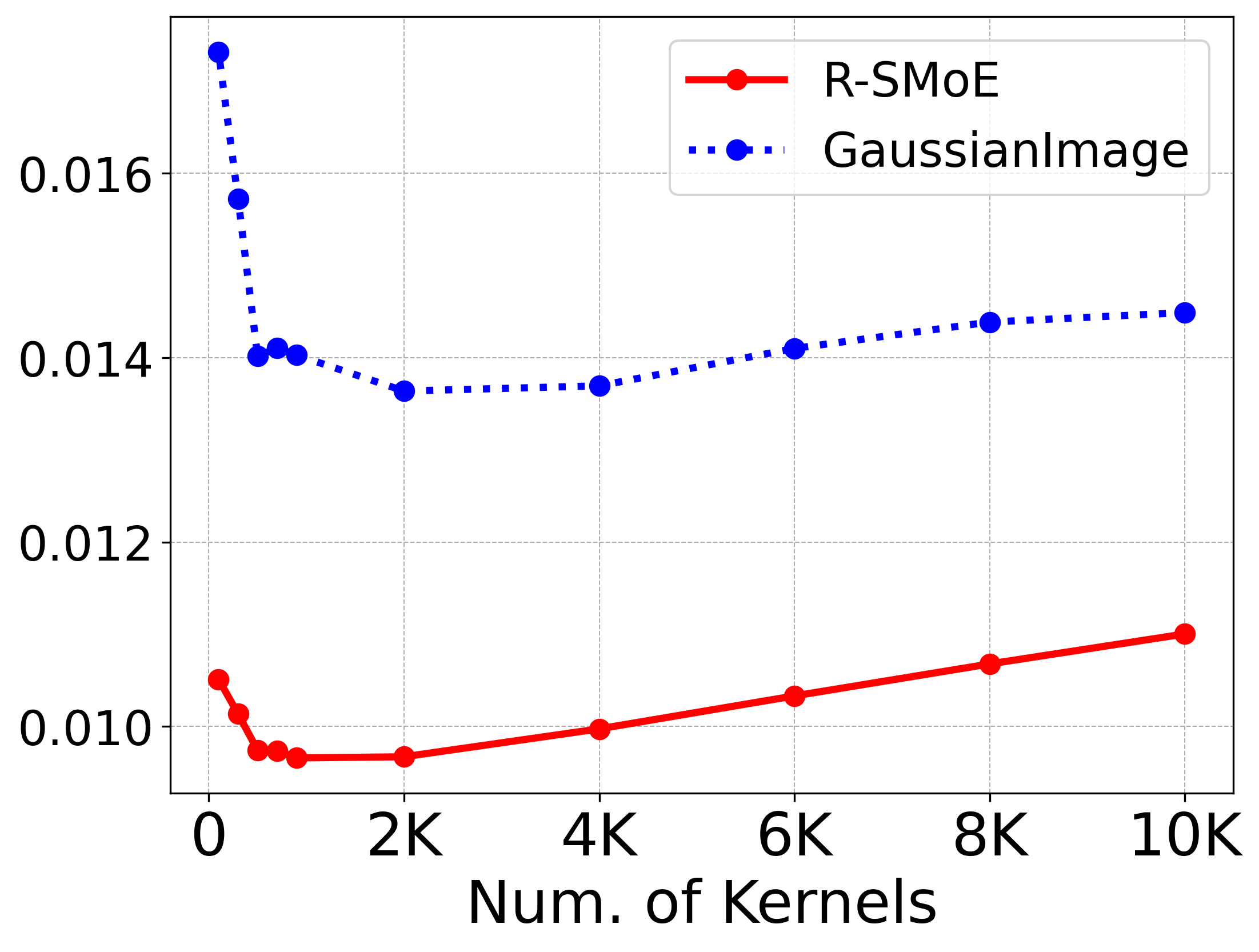}}
\vspace{0em}
\caption{Training time comparison \textcolor{black}{between R-SMoE and GSMoE highlighting significant reductions in computation time achieved by R-SMoE.}}
\vspace{-0em}
\label{fig9}
\end{figure}

Fig. \ref{fig8} presents the loss convergence comparison between the global method (GSMoE) and the rasterized method (R-SMoE) using different numbers of Gaussian kernels. Notably, as the number of Gaussian kernels increases, the convergence rate of the global method slows considerably, while the rasterized method maintains its rapid convergence, even with a larger kernel count. This behavior can be attributed to the nature of global methods, where the gradient for Gaussian parameters is accumulated across all pixels, including those with minimal contribution, leading to slower convergence. The redundant consideration of low-contributing pixels increases the time required for gradient calculation and the number of iteration\textcolor{black}{s} needed to reach the optimal solution. In contrast, the rasterized method efficiently handles the increased number of kernels without a significant slowdown in convergence.

Fig. \ref{fig9} further depicts the relationship between training time and the total number of kernels. Fig. \ref{fig9}(a) compares the training time of both global and rasterized methods. Although a higher number of Gaussians typically results in better performance, as evidenced in Fig. \ref{fig4}, R-SMoE drastically reduce\textcolor{black}{s} the optimization run-time, achieving speeds up to 1000 times faster than global SMoE. \textcolor{black}{This} is achieved by leveraging rasterization to enable localized kernel lookups and massively parallel GPU execution.

While increasing the number of kernels has some effect on the run-time of R-SMoE, this increase is minimal and does not significantly impact the method's computational speed and reconstruction quality when compared to global SMoE. 

Global methods require up to 100× more training time than rasterized methods\textcolor{black}{,} as shown in Fig.~\ref{fig9}(a). Due to this \textcolor{black}{large-scale} difference in training time, the variations among the rasterized methods are not clearly visible in the same plot. To address this, Fig. \ref{fig9}(b) zooms in on the training time of rasterized methods. At first glance, one might expect that a smaller total number of kernels would result in shorter training times, given the presumed computational simplicity. However, the observed trend reveals the opposite. This phenomenon can be explained by the nature of the training process itself. Unlike the rendering stage discussed in Fig. \ref{fig5}, where only a trained subset of kernels is selected for each block, the training phase begins with a large number of kernels per block. The kernel distribution is initially broad and gradually condenses as training progresses. When the total number of kernels is small (e.g., 100), each kernel must contribute to multiple blocks, leading to significant overlap across blocks. This overlap increases the number of kernels per block in the early stages of training, thereby requiring more time for the kernel distribution to condense. Conversely, when the total number of kernels is large, the initial kernel distribution is much denser, and each kernel is responsible for fewer blocks. This smaller overlap between kernels results in a more efficient training process, as fewer iterations are required to achieve a condensed kernel distribution. Consequently, the training time for a larger total number of kernels is shorter than that for a smaller total number of kernels.

\begin{table*}[t]
\captionsetup{justification=centering, labelsep=newline}
\centering
\caption{Quantitative evaluation of multi-model inference \textcolor{black}{performance across varying noise levels and model counts.}}
\begin{tabular}{>{\centering}m{55pt}>{\centering}m{10pt}>{\centering}m{15pt}>{\centering}m{15pt}>{\centering}m{23pt}>{\centering}m{10pt}>{\centering}m{15pt}>{\centering}m{15pt}>{\centering}m{23pt}>{\centering}m{10pt}>{\centering}m{15pt}>{\centering}m{15pt}>{\centering}m{23pt}>{\centering}m{10pt}>{\centering}m{15pt}>{\centering}m{15pt}>{\centering\arraybackslash}m{23pt}}

& \multicolumn{4}{c}{Gaussian noise $\sigma^2$ = 0.0037} & \multicolumn{4}{c}{Gaussian noise $\sigma^2$ = 0.005} & \multicolumn{4}{c}{Gaussian noise $\sigma^2$ = 0.01} & \multicolumn{4}{c}{Gaussian noise $\sigma^2$ = 0.05} \\
\hline
Method & PSNR & SSIM & LPIPS & Time(s) & PSNR & SSIM & LPIPS & Time(s) & PSNR & SSIM & LPIPS & Time(s) & PSNR & SSIM & LPIPS & Time(s)\\

MM-RSMoE-64 & \textbf{31.38} & \textbf{0.8544} & 0.2519 & 192 & \textbf{30.74} & \textbf{0.8330} & 0.2740 & 192& \textbf{28.93} & \textbf{0.7643} & \textbf{0.3343} & 192& \textbf{23.49} & \textbf{0.5191} & \textbf{0.5065} & 192\\
MM-RSMoE-16 & 31.37 & 0.8518 & \textbf{0.2505} & 48 & 30.69 & 0.8285 & \textbf{0.2738} & 48 & 28.79 & 0.7553 & 0.3373 & 48 & 23.31 & 0.5058 & 0.5110 & 48 \\
MM-RSMoE-1 & 29.64 & 0.7832 & 0.3259 & \textbf{3} & 28.86 & 0.7485 & 0.3546 & \textbf{3}& 26.80 & 0.6509 & 0.4279 & \textbf{3} & 21.25 & 0.3868 & 0.5895 & \textbf{3}\\

\end{tabular}
\label{table2}
\end{table*}

\begin{figure}[t]
\centering
\subfloat{\includegraphics[trim={0cm 0.cm 0.cm 0cm},clip,width=1\linewidth]{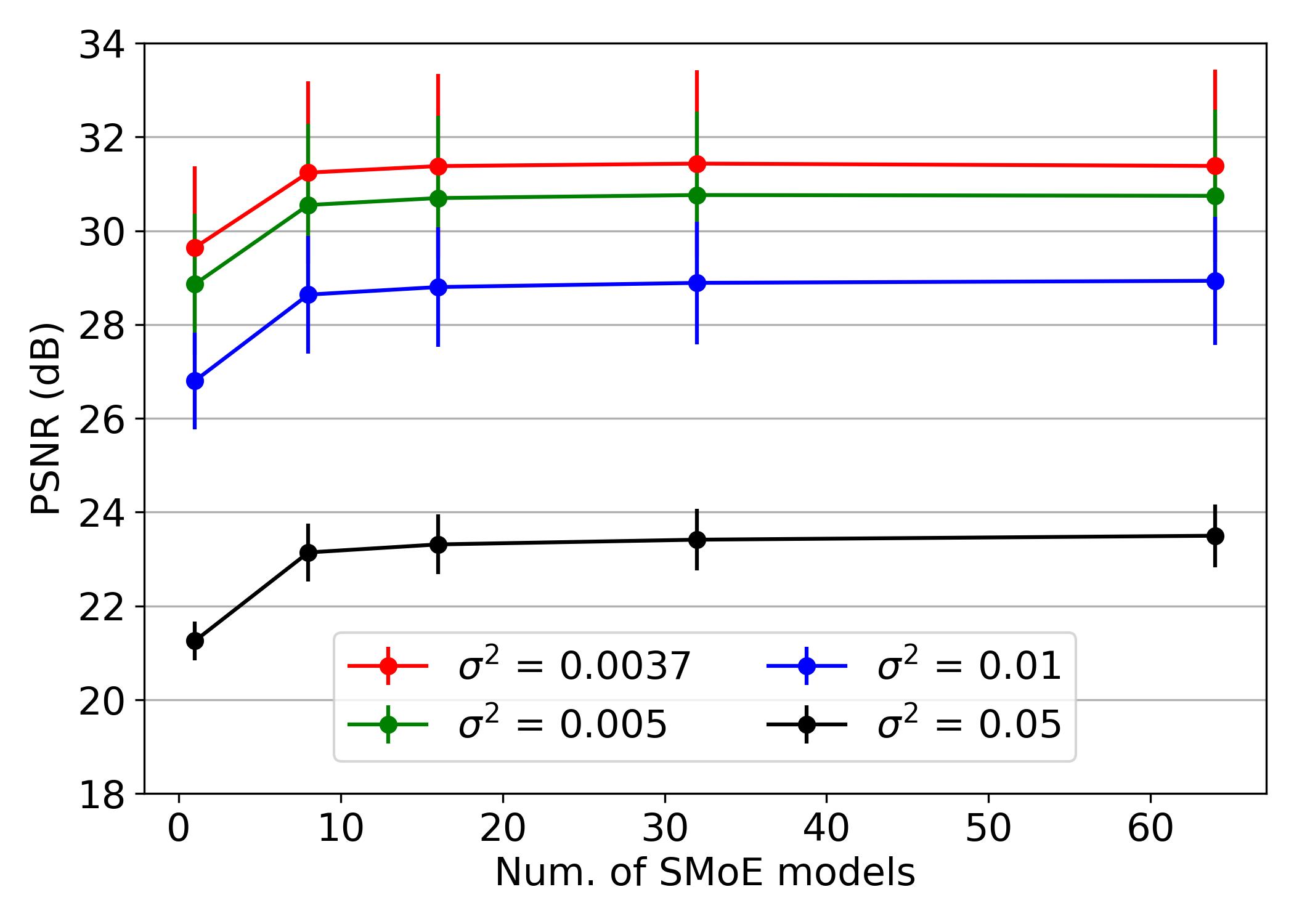}}    
\vspace{0em}
\caption{Multi-model inference performance with varying Gaussian noise variance and different numbers of SMoE hypotheses per pixel \textcolor{black}{demonstrating robustness of the proposed segmentation-guided multi-hypothesis strategy under noise conditions.}}
\vspace{-0em}
\label{fig9}
\end{figure}
\begin{figure*}[t]
\centering
\subfloat{\includegraphics[trim={0cm 0.cm 0cm 0cm},clip,width=1.\linewidth]{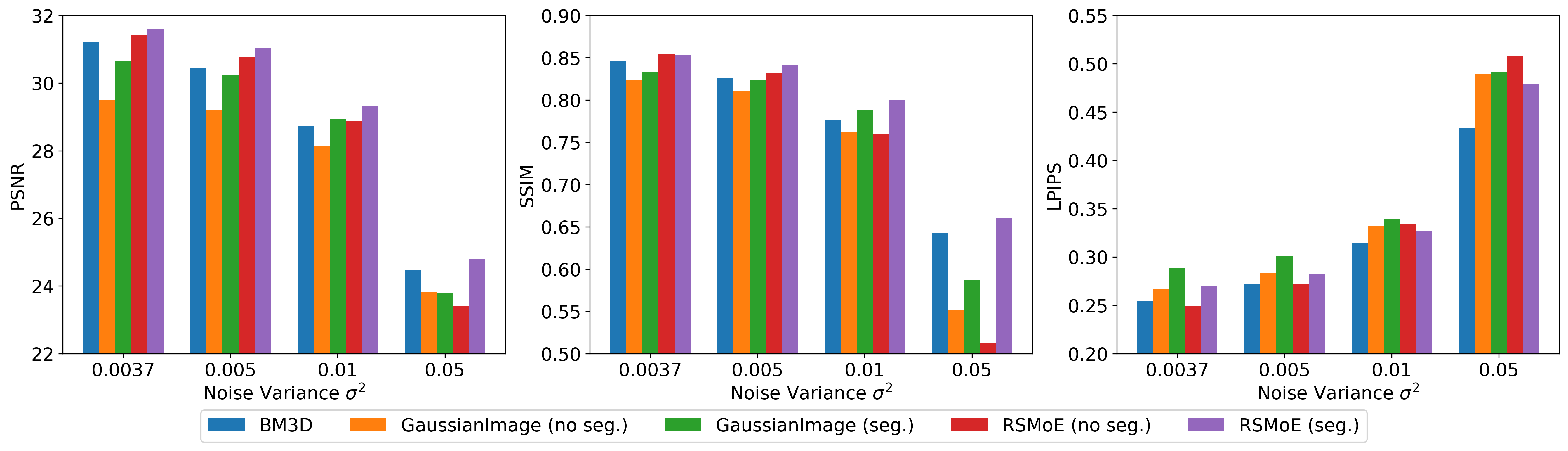}}
\vspace{0em}
\caption{Denoising performance of BM3D, GaussianImage with and without segmentation initialization, and R-SMoE with and without segmentation initialization across different Gaussian noise variances ($\sigma^2$).}
\vspace{-0em}
\label{fig10}
\end{figure*}

\subsection{Denoising}
While the prime focus of the paper is to demonstrate the improvements in image regression and acceleration for R-SMoE models, it is valuable to highlight \textcolor{black}{their} benefits in denoising. To this end, we evaluate block-overlapping multi-model (MM) inference for both R-SMoE and GaussianImage RBF regression, with and without segmentation-initialization.

Table \ref{table2} and Fig. \ref{fig9} illustrate the impact of MM inference in R-SMoE (MM-RSMoE) for denoising; see Section~V.C for details. Without leveraging multiple models, residual noise tends to persist, yielding reconstructions that barely improve upon the noisy input. By aggregating predictions from several independently trained models, MM-RSMoE effectively averages out stochastic noise—a strategy that directly reduces the second term in Eq. (\ref{eq14}), which scales inversely with the number of models $H$. More models, less noise.

Interestingly, increasing the number of models does not always lead to better denoising performance. As shown in Fig. \ref{fig9}, 16 models are sufficient to achieve a significant noise reduction, with diminishing returns beyond this point. This suggests that a moderate number of models can effectively balance denoising performance and computational efficiency. In Table \ref{table2}, the R-SMoE with multi-model inference is denoted by MM-RSMoE-*, where * represents the number of models used. Even though the computation time for MM-RSMoE scales linearly with the number of models, the training of each model can be launched in parallel. This means that the overall denoising time of MM-RSMoE can be significantly reduced when utilizing multiple GPUs, further enhancing its practicality for large-scale applications.

Table \ref{table3} presents a comprehensive comparison of MM-RSMoE (8 and 64 models) with and without segmentation-based initialization, evaluated across various noise levels ($\sigma^2$ = [0.0037, 0.005, 0.01, 0.05]). Results without segmentation use random initialization, while segmentation-enhanced variants are denoted as “Method-seg-*”, where the asterisk indicates the segmentation threshold used during initialization. These thresholds control the similarity criterion between segments—lower values enforce stricter homogeneity. In our experiments, we evaluate thresholds of 10 and 20 to test robustness to this parameter.

When the MM-RSMoE is initialized without segmentation, using a random initialization, it achieves slightly better results than BM3D for images with low noise variance. However, as the noise variance increases, the performance of the MM-RSMoE declines relative to BM3D, as shown in Table \ref{table3}. This reduction in performance can be attributed to the tendency of \textcolor{black}{random} initialization. In addition, SMoE-based methods are excellent at preserving high-frequency details indiscriminately, including the noise. Consequently, even with the multi-model inference, the well-preserved noise disrupts the continuity and homogeneity of the image, leading to suboptimal results, particularly in scenarios with high noise variance.


The segmentation-based initialization strategy allocates kernels adaptively—fewer in flat regions, more in texture-rich areas. This smarter kernel distribution avoids overfitting to noise in homogeneous regions, resulting in cleaner, more coherent reconstructions. As discussed in Section 5.B, Eq.~(\ref{eq14}) reveals that the residual noise variance scales inversely with $M$, the number of pixels covered by a kernel. By reducing kernel density in flat areas through segmentation, we effectively enlarge each kernel’s support, increasing $M$ and thus suppressing noise more efficiently. This directly explains the observed gains in denoising performance. As shown in Table \ref{table3} and Fig. \ref{fig10}, the segmentation-enhanced models consistently outperform BM3D across various noise levels ($\sigma^2$ = [0.0037, 0.005, 0.01, 0.05]). Specifically, MM-RSMoE-seg-* achieves a PSNR gain of ~0.3 dB and an SSIM improvement of 0.2 over BM3D, demonstrating the efficacy of segmentation-based initialization.

Furthermore, the results indicate that denoising performance is largely insensitive to the specific segmentation threshold—both threshold values yield similar improvements, suggesting robustness to segmentation granularity. Table \ref{table3} also illustrates that simple block-overlapping averaging of 8 hypotheses per pixel already provides excellent denoising results. Compared to 64 models per pixel the processing time is greatly reduced. Denoising with block-overlapping GaussianImage regression is also possible, but results are not competitive with R-SMoE and BM3D.

\begin{table*}[h!]
\captionsetup{justification=centering, labelsep=newline}
\centering
\caption{Quantitative evaluation for denoising \textcolor{black}{performance across different methods and noise level.}}
\begin{tabular}{>{\centering}m{2pt}>{\centering}m{74pt}>{\centering}m{10pt}>{\centering}m{15pt}>{\centering}m{15pt}>{\centering}m{15pt}>{\centering}m{10pt}>{\centering}m{15pt}>{\centering}m{15pt}>{\centering}m{15pt}>{\centering}m{10pt}>{\centering}m{15pt}>{\centering}m{15pt}>{\centering}m{15pt}>{\centering}m{10pt}>{\centering}m{15pt}>{\centering}m{15pt}>{\centering\arraybackslash}m{15pt}}

&& \multicolumn{4}{c}{Gaussian noise $\sigma^2$ = 0.0037} & \multicolumn{4}{c}{Gaussian noise $\sigma^2$ = 0.005} & \multicolumn{4}{c}{Gaussian noise $\sigma^2$ = 0.01} & \multicolumn{4}{c}{Gaussian noise $\sigma^2$ = 0.05} \\
\hline
\multicolumn{2}{c}{Method} & PSNR & SSIM & LPIPS & Time(s) & PSNR & SSIM & LPIPS & Time(s) & PSNR & SSIM & LPIPS & Time(s) & PSNR & SSIM & LPIPS & Time(s)\\

\multicolumn{2}{c}{Noisy}& 24.47 & 0.5063 & 0.2622 & n/a& 23.19 & 0.4517 & 0.3174 & n/a & 20.28 & 0.3352 & 0.4675 & n/a	& 13.90	& 0.1423 & 0.8791 & n/a\\
\multicolumn{2}{c}{BM3D} & 31.23 & 0.8462 &	0.2544 & \textbf{4}	& 30.46	& 0.8263 & \textbf{0.2726} & \textbf{4}   & 28.74 & 0.7764 & \textbf{0.3144} & \textbf{4}   & 24.48 & 0.6424 & \textbf{0.4339} & \textbf{4}\\
\hline
\multirow{6}{*}{\rotatebox{90}{64 models}}
& GaussianImage        & 29.53 & 0.8259 &	0.2659 & 640& 29.21 & 0.8120 & 0.2826 & 640	& 28.18 & 0.7641 & 0.3311 & 640	& 23.87 & 0.5542 & 0.4874 & 640\\
& GaussianImage-seg-20 & 29.24 & 0.8089 &	0.3057 & 640& 29.51 & 0.8070 & 0.3186 & 640 & 28.53 & 0.7749 & 0.3526 & 640 & 24.16 & 0.6029 & 0.4793 & 640\\
& GaussianImage-seg-10 & 30.71 & 0.8357 &	0.2859 & 640& 30.30 & 0.8265 & 0.2985 & 640	& 29.00	& 0.7908 & 0.3356 & 640 & 23.84 & 0.5891 & 0.4888 & 640\\
& MM-RSMoE             & 31.38 & 0.8544 & \textbf{0.2519} & 192& 30.74	& 0.8330 & 0.2740 & 192 & 28.93	& 0.7643 & 0.3343 & 192 & 23.50 & 0.5191 & 0.5065 & 192\\
& MM-RSMoE-seg-20      & \textbf{31.73} & \textbf{0.8553} & 0.2765 & 192& \textbf{31.21}	& \textbf{0.8460} & 0.2872 & 192 & \textbf{29.60} & \textbf{0.8111} & 0.3264 & 192 & 24.31 & 0.6565 & 0.4712 & 192 \\
& MM-RSMoE-seg-10         & 31.61 & 0.8535 &	0.2707 & 192& 31.05 & 0.8421 & 0.2830 & 192 & 29.40 & 0.8023 & 0.3261 & 192 & \textbf{24.94} & \textbf{0.6709} & 0.4775 & 192\\
\hline
\multirow{6}{*}{\rotatebox{90}{8 models}}
& GaussianImage        & 29.29 & 0.8111 & 0.2766 & 80 & 28.98 & 0.7965 & 0.2936 & 80 & 27.93 & 0.7465 & 0.3434 & 80 & 23.61 & 0.5344 & 0.5017& 80 \\
& GaussianImage-seg-20 & 28.87 & 0.7890 & 0.3244 & 80 & 29.17 & 0.7904 & 0.3372 & 80 & 28.21 & 0.7573 & 0.3699 & 80 & 23.90 & 0.5831 & 0.4926& 80 \\
& GaussianImage-seg-10 & 30.43 & 0.8247 & 0.2975 & 80 & 30.02 & 0.8152 & 0.3098 & 80 & 28.72 & 0.7778 & 0.3495 & 80 & 23.58 & 0.5720 & 0.5017& 80 \\
& MM-RSMoE             & 31.23 & 0.8469 & 0.2541 & 24 & 30.54 & 0.8226 & 0.2782 & 24 & 28.63 & 0.7470 & 0.3431 & 24 & 23.14 & 0.4941 & 0.5174& 24 \\
& MM-RSMoE-seg-20	   & 31.60 & 0.8523 & 0.2771 & 24  & 30.59 & 0.8263 & 0.3027 & 24 & 29.09 & 0.7878 & 0.3437 & 24 & 24.73 & 0.6566 & 0.4905 & 24 \\
& MM-RSMoE-seg-10	   & 31.44 & 0.8494 & 0.2711 & 24  & 30.85 & 0.8365 & 0.2860 & 24 & 29.10 & 0.7913 & 0.3330 & 24 & 24.66 & 0.6530 & 0.4850 & 24 \\
\end{tabular}
\label{table3}
\end{table*}

\begin{figure*}[h!]
    \centering
    \hfill
    \begin{minipage}{1\linewidth}  
        \includegraphics[width=\linewidth]{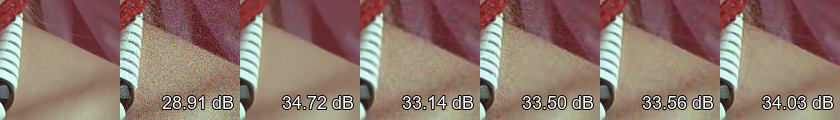}
        \includegraphics[width=\linewidth]{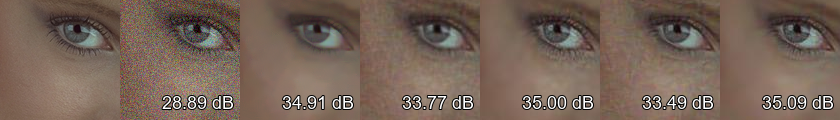}
    \end{minipage}
    \\
    \hfill
    \begin{minipage}{1\linewidth}
        \includegraphics[width=\linewidth]{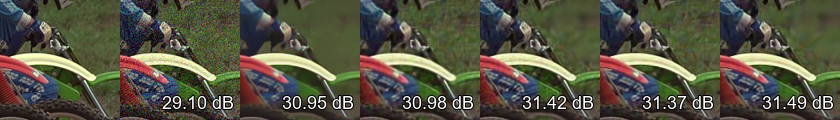}
        \includegraphics[width=\linewidth]{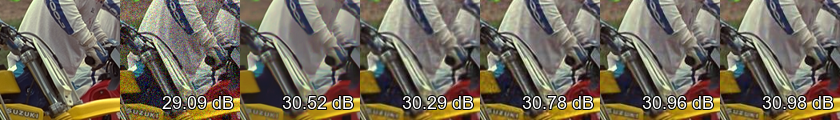}
    \end{minipage}
    \\
    \hfill
    \begin{minipage}{1\linewidth}
        \includegraphics[width=\linewidth]{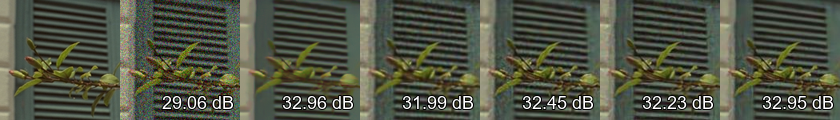}
        \includegraphics[width=\linewidth]{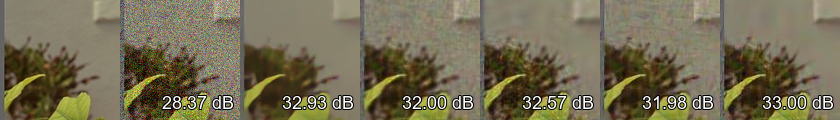}
    \end{minipage}
    \\
    \hfill
    \begin{minipage}{1\linewidth}
        \includegraphics[width=\linewidth]{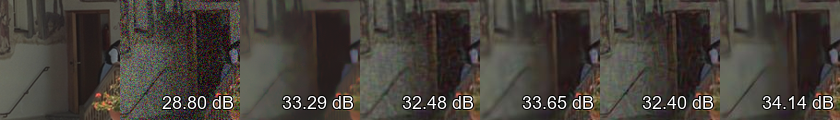}
        \includegraphics[width=\linewidth]{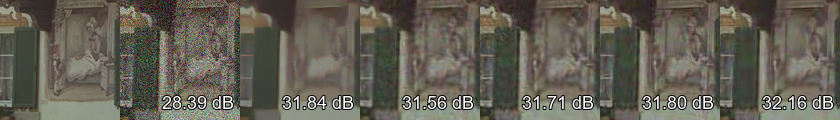}
    \end{minipage}
    \\
    \vspace{1pt}
    \begin{tabular}{>{\centering}p{0.12\linewidth}>{\centering}p{0.12\linewidth}>{\centering}p{0.12\linewidth}>{\centering}p{0.11\linewidth}>{\centering}p{0.14\linewidth}>{\centering}p{0.105\linewidth}>{\centering}p{0.13\linewidth}}
    Original & Noisy & BM3D & \shortstack{GaussianImage} & \shortstack{GaussianImage-seg} & \shortstack{MM-RSMoE} & \shortstack{MM-RSMoE-seg}
    \end{tabular}

    \caption{\textcolor{black}{Visualization of denoising results for BM3D, GaussianImage, and the proposed R-SMoE with a noise variance of 0.01. \textit{Method-seg} in this figure refers to “Method-seg-10”.}}
    \label{fig11}
\end{figure*}

In Fig. \ref{fig11}, the visual comparison between the proposed MM-RSMoE and the BM3D method highlights the superior performance of MM-RSMoE in preserving high-frequency details. While BM3D tends to smooth out these details, resulting in a loss of texture and edge sharpness, MM-RSMoE effectively retains them, ensuring that the image remains sharp and well-defined. However, without the integration of segmentation-based initialization, MM-RSMoE tends to reconstruct noise in flat regions, which negatively impacts overall image quality. By incorporating segmentation-based initialization, MM-RSMoE-seg can effectively distinguish between textured areas and flat regions, leading to improved noise suppression in smoother regions. This enhancement is particularly noticeable in the flat areas of the visualized results, where segmentation initialization reduces noise and produces a cleaner, more visually appealing image. For example, in the flower image (Fig. \ref{fig11}), MM-RSMoE-seg suppresses noise in the flat background while preserving the flower’s intricate details, highlighting its dual talent for denoising and structure preservation.

\begin{figure*}[t]
\begin{tabular}{cc}
\raisebox{1.0\height}{\rotatebox{90}{wo. sharpen}} &
\subfloat{\includegraphics[trim={0cm 0cm 0cm 0cm},clip,width=0.16\linewidth]{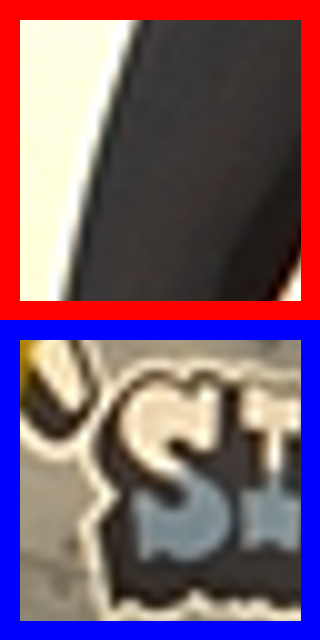}}
\subfloat{\includegraphics[trim={0cm 0cm 0cm 0cm},clip,width=0.16\linewidth]{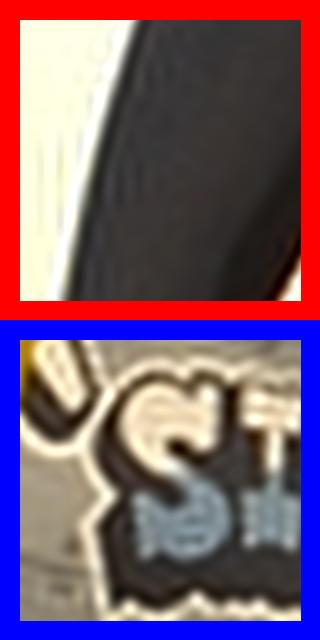}}
\subfloat{\includegraphics[trim={0cm 0cm 0cm 0cm},clip,width=0.16\linewidth]{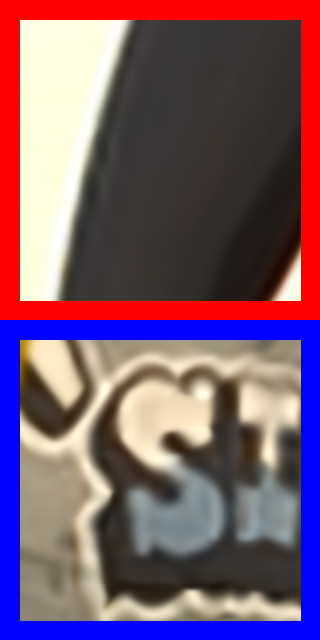}}
\subfloat{\includegraphics[trim={0cm 0cm 0cm 0cm},clip,width=0.16\linewidth]{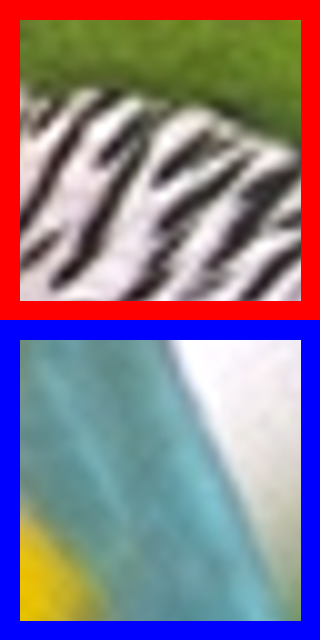}}
\subfloat{\includegraphics[trim={0cm 0cm 0cm 0cm},clip,width=0.16\linewidth]{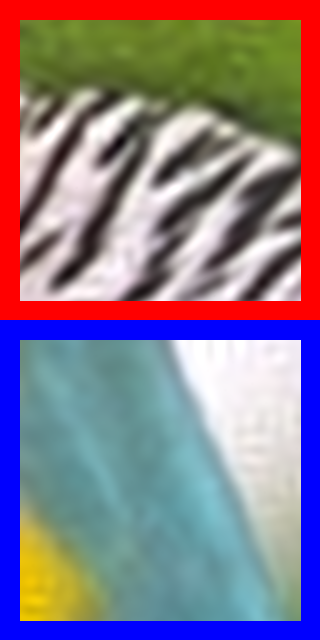}}
\subfloat{\includegraphics[trim={0cm 0cm 0cm 0cm},clip,width=0.16\linewidth]{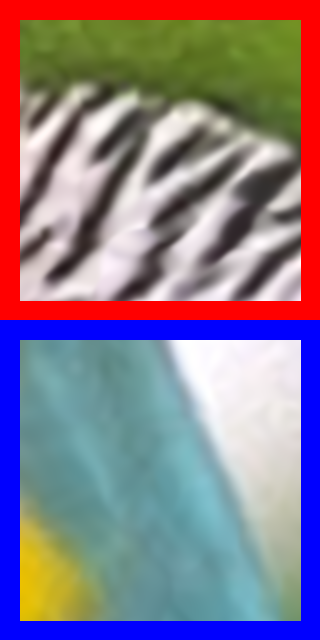}}

\end{tabular}
\\
\begin{tabular}{cc}
\raisebox{0.9\height}{\rotatebox{90}{with sharpen}} &
\subfloat{\includegraphics[trim={0cm 0cm 0cm 0cm},clip,width=0.16\linewidth]{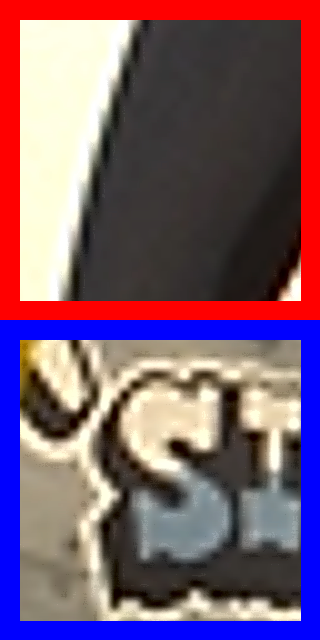}}
\subfloat{\includegraphics[trim={0cm 0cm 0cm 0cm},clip,width=0.16\linewidth]{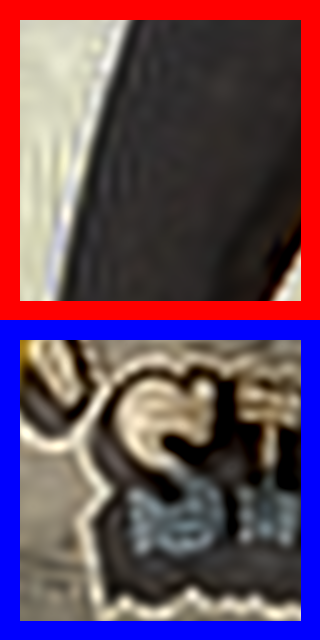}}
\subfloat{\includegraphics[trim={0cm 0cm 0cm 0cm},clip,width=0.16\linewidth]{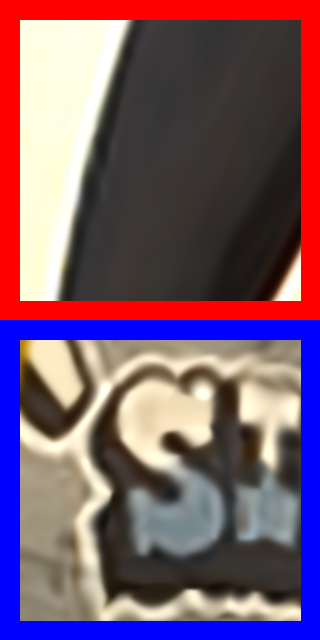}}
\subfloat{\includegraphics[trim={0cm 0cm 0cm 0cm},clip,width=0.16\linewidth]{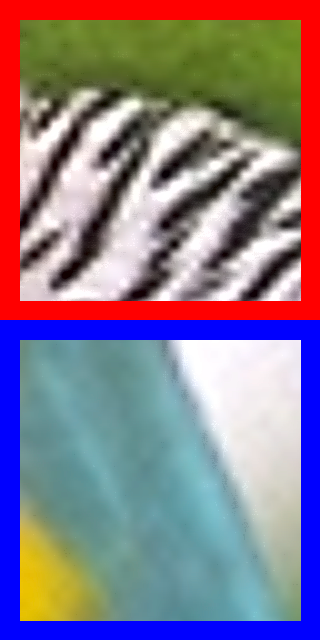}}
\subfloat{\includegraphics[trim={0cm 0cm 0cm 0cm},clip,width=0.16\linewidth]{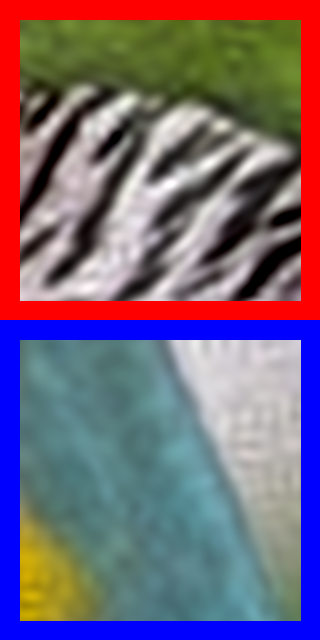}}
\subfloat{\includegraphics[trim={0cm 0cm 0cm 0cm},clip,width=0.16\linewidth]{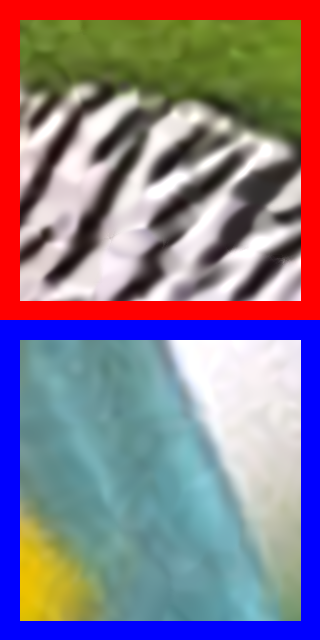}}
\end{tabular}
\\
\begin{minipage}[b]{0.95\linewidth}
    \begin{tabular}{w{c}{5mm}w{c}{25mm}w{c}{24mm}w{c}{24mm}w{c}{25mm}w{c}{25mm}w{c}{25mm}} 
            &Bicubic & GaussianImage \cite{zhang2024gaussianimage} & R-SMoE (Ours)&Bicubic & GaussianImage \cite{zhang2024gaussianimage} & R-SMoE (Ours)
    \end{tabular}
\end{minipage}
\\
  
\vspace{0em}
\caption{Comparison of super-resolution results with and without the sharpening factor at a magnification factor of 10x.}
\vspace{-0em}
\label{fig13}
\end{figure*}
\subsection{Native Super-Resolution and Sharpening}
R-SMoE and GS-RBF (implemented in GaussianImage) regression allow native sharpening and super-resolution of images using kernel manipulation. As previously introduced in Section II.B, this is achieved by scaling the kernel bandwidths with a sharpening factor and resampling the continuous regression function to any scale and/or pixel raster. Fig. \ref{fig13} provides visual results on two crops of Kodak propeller plane image with $10\times$ magnification, with and without sharpening. As a reference, bicubic interpolation provides the expected staircase artifact due to the separable filter design employed. Subsequent sharpening attenuates this effect. Both GS-RBF and R-SMoE employ Gaussians that steer along edges, which avoids these artifacts. 

As already discussed with Fig. \ref{fig2}, GaussianImage results in overshoot ringing artifacts near edges at low resolution. In super-resolution, this is more clearly visible (Fig. \ref{fig13}) and drastically attenuated with sharpening. With R-SMoE no such artifacts appear at low resolution and also not after magnification -- neither without nor with sharpening. 

It appears that the results of the simple 1D experiment discussed in Fig. \ref{fig2}(b) are sufficient to explain the drastic differences in quality between GaussianImage and R-SMoE for superresolution and sharpening on real 2D imagery.

\begin{figure}[!t]
    \centering
    \subfloat[Ground Truth]{\includegraphics[width=0.5\linewidth]{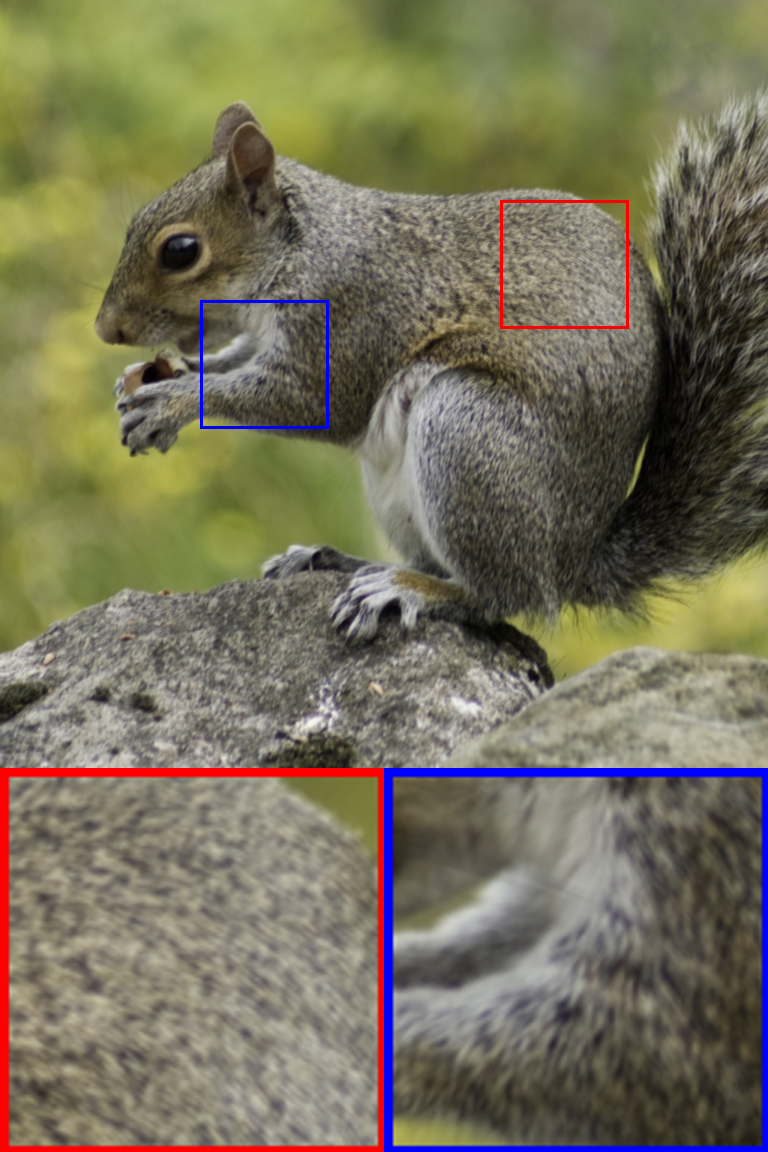}}
    \hfil
    \subfloat[R-SMoE]{\includegraphics[width=0.5\linewidth]{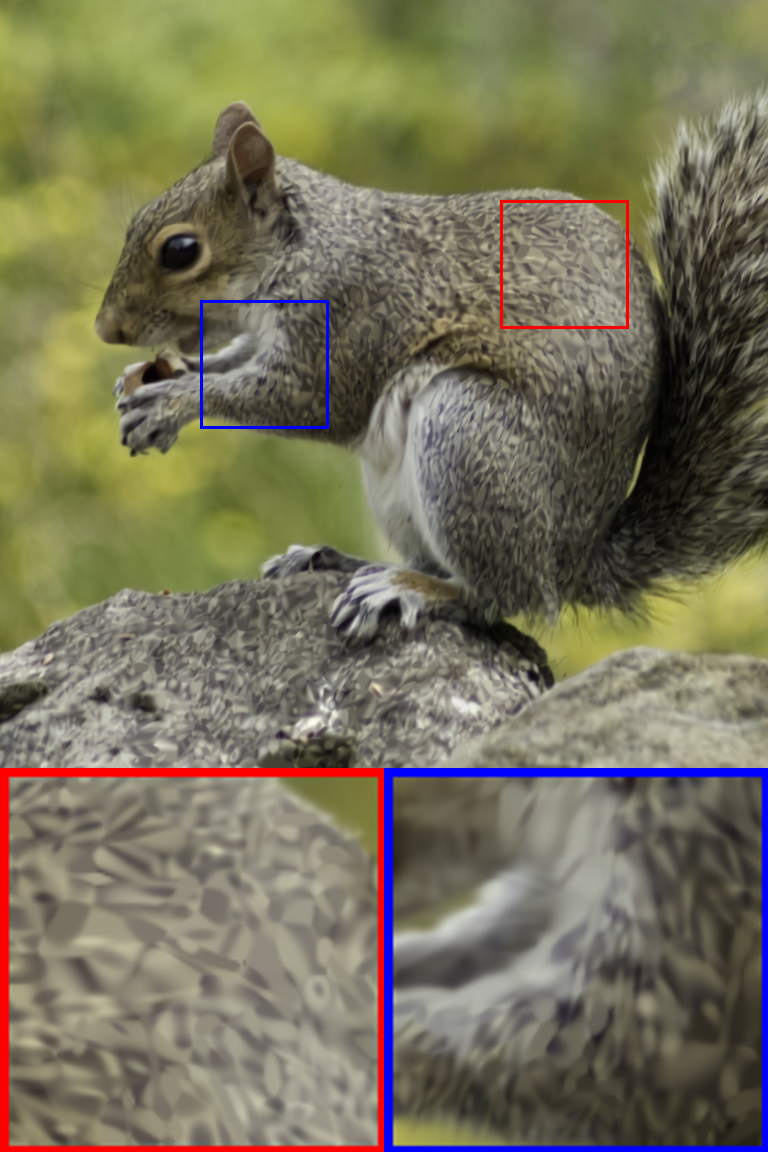}}
    \caption{\textcolor{black}{\textbf{Illustration of failure cases in high-frequency textures. 
    Example from the DIV2K test set showing reconstruction of fine fur details. 
    R-SMoE preserves sharp structural boundaries but tends to segment highly stochastic textures into smoother, coherent patches, leading to a loss of micro-level details.}}}
    \label{fig14}
\end{figure}

\textcolor{black}{\section{Discussion and Limitations}}
\textcolor{black}{
The proposed R-SMoE model delivers marked improvements in reconstruction quality and computational efficiency. However, R-SMoE exhibits limitations when applied to extremely high-frequency, stochastic textures—e.g. fur, grass, or tightly woven fabrics. Its gating mechanism, rooted in edge-aware segmentation, excels at preserving crisp structural boundaries but inevitably simplifies highly irregular regions into broader, smoother patches. This segmentation-like bias manifests as oversimplified textures, as demonstrated in Fig.~\ref{fig14}, where fine fur details lose their complexity compared to the ground truth.}
\textcolor{black}{
Though R-SMoE reduces encoding time significantly relative to prior art (Tables \ref{table1} and \ref{table2}), absolute encoding remains on the order of one to two minutes for high-resolution images. Real-time or near-real-time decoding is within reach, but encoding speed still hinders strict real-time applications demanding instantaneous bidirectional processing.} 

\textcolor{black}{While our experiments focused on benchmark datasets such as Kodak and DIV2K, we acknowledge that broader validation on domain-specific data (e.g., medical, satellite, or environmental imagery) would further demonstrate the robustness of R-SMoE. Extending our method to these datasets may require additional refinements, particularly to adapt to unique structural patterns and noise characteristics. Similarly, although our study considered standard RGB images, the method is conceptually extendable to multi-channel or multi-spectral data, and it can scale to higher-resolution images (e.g., 4K) with appropriate computational resources. We consider these directions important avenues for future work.}

\textcolor{black}{Another relevant factor is the choice of rasterization granularity. In principle, finer granularity may improve accuracy by allowing kernels to adapt more locally, while coarser granularity reduces computational cost. In the current design, kernel sizes already span multiple blocks, which implicitly balances these trade-offs. A dedicated ablation study isolating rasterization granularity would provide further insight, but this analysis is left for future work.}

\textcolor{black}{
Addressing these limitations calls for adaptive kernel sampling and progressive encoding strategies—tools to better capture stochastic textures and trim encoding latency without sacrificing fidelity.}

\textcolor{black}{
\textbf{Scope Clarification:} This work confines itself to 2D image regression. Extending rasterized SMoE to 3D Gaussian splatting entails new complexities—depth consistency, view-dependent gating—that fall outside this paper’s scope. These challenges require a dedicated investigation.}

\section{Summary and Conclusion}

In this paper we introduced Rasterized Steered Mixture of Experts (R-SMoE) as a sparse, fast, and memory-efficient framework for 2D image regression. Compared to previous ``global'' optimization strategies, \textcolor{black}{training and reconstruction run-times improve drastically}. \textcolor{black}{Against} Rasterized Gaussian Splatting, R-SMoE \textcolor{black}{delivers} significant \textcolor{black}{gains in both} quality and speed.

We provided in-depth insight into the similarities and differences between Gaussian Splatting and SMoE regression. Both methods share significant conceptual similarities. However, the edge-aware soft-gating network strategy of SMoE is \textcolor{black}{fundamentally} different \textcolor{black}{from} Radial Basis Function for Gaussian Splatting and provides significantly sparser models. This makes SMoE regression very attractive beyond mere high-quality image reconstruction, \textcolor{black}{rendering it well-suited} for applications like image denoising and super-resolution \textcolor{black}{by straightforward sharpening of the kernels}. Gaussian Splatting kernel regression, on the other hand, \textcolor{black}{demonstrates} limited capability for recovering images from noise and image magnification with \textcolor{black}{\textit{direct}} kernel editing. Experimental results confirm the superior performance \textcolor{black}{of} SMoE regression. 

SMoEs have already demonstrated excellent results in 3D image and video reconstruction and compression, as well as in 4D/5D light field representation and coding. The extension of Rasterized 2D SMoE modeling to higher-dimensional data \textcolor{black}{is promising, including} applications like 3D/4D noise reduction and super-resolution. While steered Gaussian kernels dominate these models, strong performance has also been reported using steered Epanechnikov kernels, particularly for image and light field data. Based on the results of Rasterized SMoEs in this paper, it is expected that these representations can be optimized and reconstructed with excellent quality and speed-ups \textcolor{black}{of} orders of magnitude. 

\textcolor{black}{Nevertheless, the performance gains reported here should be interpreted within the method’s intended scope: R-SMoE excels in fast, edge-aware reconstruction but may underrepresent extremely high-frequency details (e.g., fur or dense textures) and still requires several minutes per image for encoding, limiting its real-time applicability. These trade-offs position R-SMoE as a practical, task-specific solution rather than a universal replacement for all Gaussian-based regression methods.}

\bibliographystyle{IEEEtran}
\bibliography{refs}

\begin{IEEEbiography}[{\includegraphics[width=1in,height=1.25in,clip,keepaspectratio]{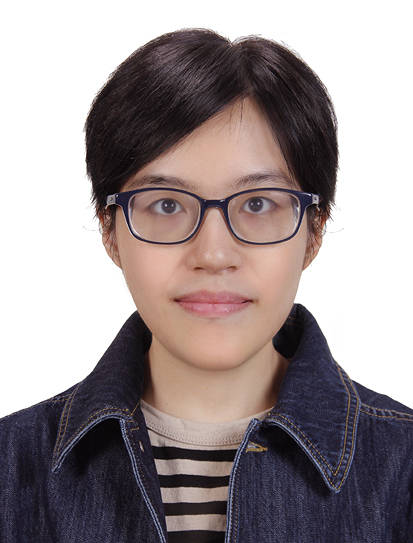}}]{Yi-Hsin Li}
received a Bachelor's degree in Electrical Engineering from National Chiao Tung University in 2018 and a Master's degree in Electrical Engineering from National Taiwan University in 2020. In November 2021, she started a double-degree Ph.D. program at the Technical University of Berlin, Germany, with a secondment to Mid Sweden University. She is in her fourth year of Ph.D. research on high-dimensional data compression, focusing on gating networks. 
\end{IEEEbiography}
\begin{IEEEbiography}[{\includegraphics[width=1in,height=1.25in,clip,keepaspectratio]{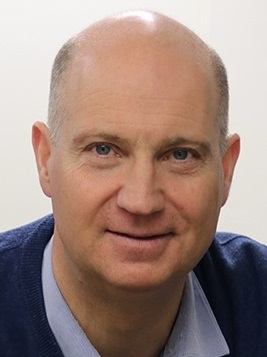}}]{Mårten~Sjöström}
(M'95, SM'17) received the M.Sc. degree from Linköping University (1992), the Licentiate of Technology degree from the Royal Institute of Technology Stockholm (1998), and the Ph.D. degree from the École Polytechnique Fédérale de Lausanne (2001). He was with ABB (1993-1994) and with CERN (1994-1996), involved with projects on signal processing. In 2001, he joined Mid Sweden University and was appointed Associate Professor (2008) and Full Professor of Signal Processing (2013). He is head of research education in Computer and System Science (2013-) and Computer Engineering (2020-). He is head and founder of the Realistic 3-D Research Group (2007-). He has served as Associate editor for IEEE Transactions on Image Processing (2022-), and for SPIE Journal of Electronic Imaging (2018-2022). He is a board member of High Performance Computing Centre North (2019-). His current research interests include Visual AI, machine learning for multidimensional signal processing and imaging, and system modeling and identification.
\end{IEEEbiography}
\begin{IEEEbiography}[{\includegraphics[width=1in,height=1.25in,clip,keepaspectratio]{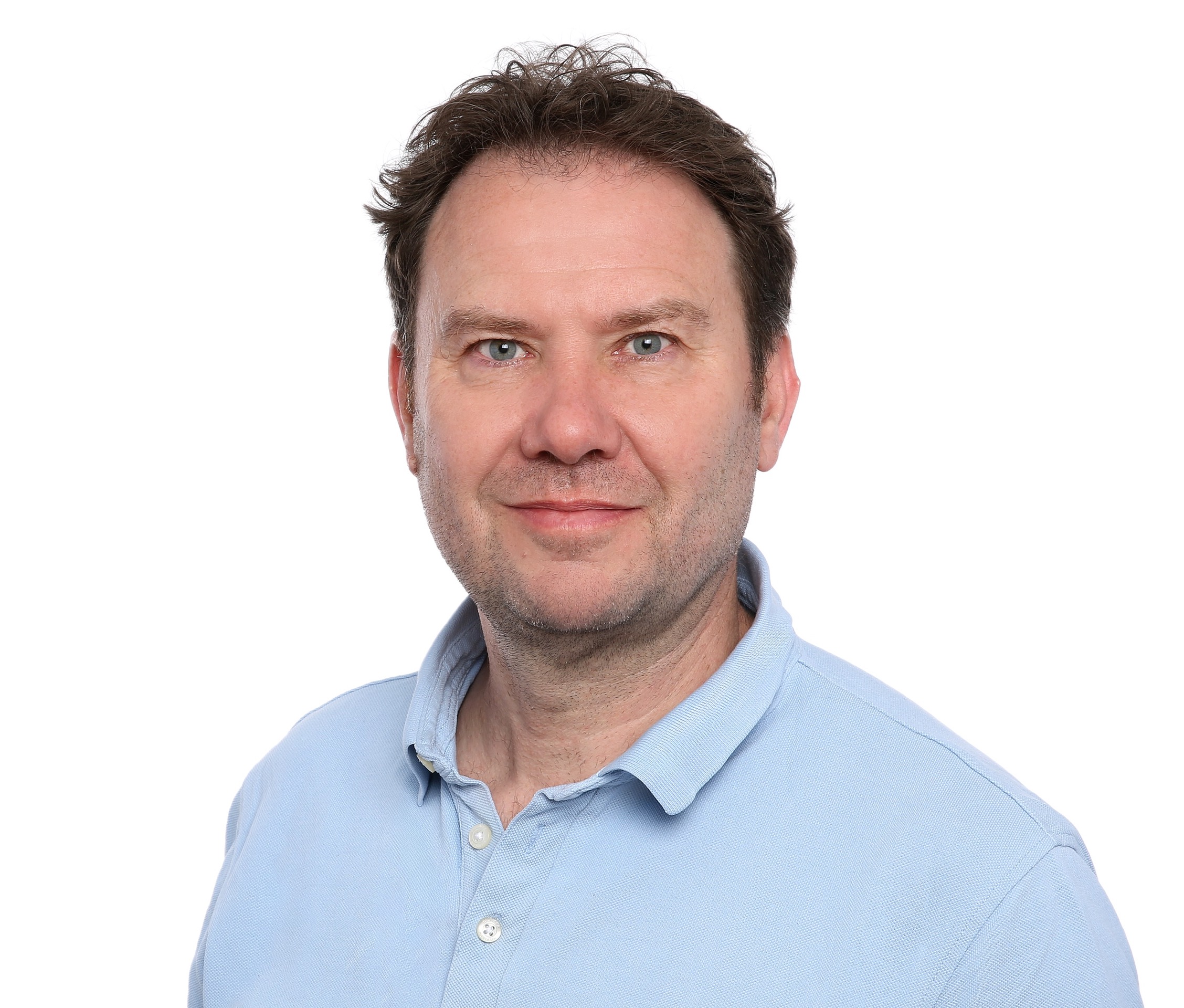}}]{Sebastian~Knorr}
(M'11, SM'19) received the Diploma and Ph.D. degree in Electrical Engineering from the Technical University of Berlin, Germany in 2002 and 2008, respectively. Between 2009 and 2016, he was the CEO/CTO of imcube labs GmbH, Germany. He was Senior Research Scientist and Lecturer at Trinity College Dublin and TU Berlin between 2017 and 2020, respectively. From 2020 to 2024, he was Full Professor at the Ernst-Abbe University of Applied Sciences Jena and is currently Full Professor for Visual Computing at the School of Computing, Communication and Business, HTW Berlin. From 2019 to 2023, he was Associate Editor of the IEEE Transactions of Multimedia, and since 2023, he is Associate Editor of the IEEE Transactions on Image Processing. His research interests include light field imaging, neural rendering, free-viewpoint-video, 3D image processing and 360$^\circ$ video.
\end{IEEEbiography}
\begin{IEEEbiography}[{\includegraphics[width=1in,height=1.25in,clip,keepaspectratio]{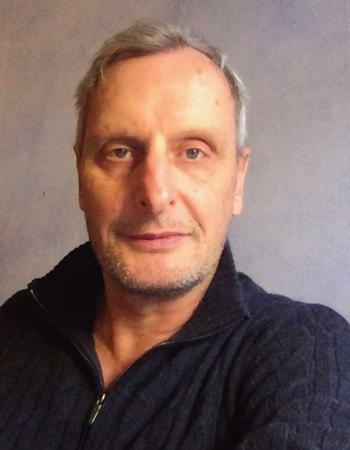}}]{Thomas~Sikora} 
(Senior Member, IEEE) is the Director of the Communication Systems Lab at Technische Universität Berlin, Germany. He received the Dipl.-Ing. and Dr.-Ing. degrees in electrical engineering from Bremen University, Bremen, Germany, in 1985 and 1989, respectively. In 1990, he joined Siemens Ltd. and Monash University, Melbourne, Australia, as a Project Leader responsible for video compression research activities in the Australian Universal Broadband Video Codec consortium.
Between 1994 and 2001, he was the Director of the Interactive Media Department at the Heinrich Hertz Institute (HHI) Berlin GmbH, Germany. In 2002, he was appointed Full Professor at TU Berlin.
Prof. Sikora has been involved in international ITU and ISO standardization activities, as well as in several European research projects, for many years. As Chairman of the ISO-MPEG (Moving Picture Experts Group) video group, he was responsible for the development and standardization of the MPEG-4 and MPEG-7 video algorithms. He is the recipient of an Engineering Emmy Award and the 1996 German ITG Award.
He has served as Associate Editor on the editorial boards of several journals, including EURASIP Signal Processing: Image Communication. From 1998 to 2002, he was an Associate Editor of the IEEE Signal Processing Magazine. He is also a past Editor-in-Chief of the IEEE Transactions on Circuits and Systems for Video Technology.
\end{IEEEbiography}

\end{document}